\documentclass[letterpaper]{article} 
\usepackage{aaai2026}   
\usepackage{times}  
\usepackage{helvet}  
\usepackage{courier}  
\usepackage[hyphens]{url}  
\usepackage{graphicx} 
\usepackage{natbib}  
\usepackage{caption} 
\nocopyright
\usepackage{algorithm}
\usepackage{algorithmic}
\usepackage{booktabs}
\usepackage{multirow}
\usepackage{xcolor}         
\usepackage{amsfonts}
\usepackage{colortbl} 
\usepackage[most]{tcolorbox}
\usepackage{utfsym}
\usepackage{newfloat}
\usepackage{listings}
\DeclareCaptionStyle{ruled}{labelfont=normalfont,labelsep=colon,strut=off} 
\floatstyle{ruled}
\newfloat{listing}{tb}{lst}{}
\floatname{listing}{Listing}
\title{Test-time Prompt Intervention}
\author{
    Chenxu Yang\textsuperscript{\rm 1,2}\thanks{\ \ \ Equal contribution. }, Qingyi Si\textsuperscript{\rm 3}\footnotemark[1], Muzhi Dai\textsuperscript{\rm 3}\footnotemark[1], Dingyu Yao\textsuperscript{\rm 1,2}, \\
    {\bf Mingyu Zheng\textsuperscript{\rm 1,2}, Minghui Chen\textsuperscript{\rm 1,2}, Zheng Lin\textsuperscript{\rm 1,2}\thanks{\ \ \ Zheng Lin is the corresponding author. }, Weiping Wang\textsuperscript{\rm 1,2}
}}
\affiliations{
   \textsuperscript{\rm 1}Institute of Information Engineering, Chinese Academy of Sciences, Beijing, China \\
  \textsuperscript{\rm 2}School of Cyber Security, University of Chinese Academy of Sciences, Beijing, China \\
  \textsuperscript{\rm 3}Huawei Technologies Co., Ltd.\\
  \textrm{\{}yangchenxu,linzheng\textrm{\}}@iie.ac.cn; 
  {siqingyi@huawei.com}
}

\usepackage{bibentry}

\begin{document}

\maketitle

\begin{abstract}
Test-time compute has led to remarkable success in the large language model (LLM) community, particularly for complex tasks, where longer chains of thought (CoTs) are generated to enhance reasoning capabilities. However, growing evidence reveals that such reasoning models often produce CoTs plagued by excessive redundancy, including repetitive verification steps and unnecessary reasoning shifts. The root cause lies in post-training of them that overly rely on outcome reward paradigms, as the data of process reward paradigms, which regulate intermediate reasoning steps, is difficult to construct at scale. 
To address this, we propose \textbf{PI} ($\pi$), a novel framework for Test-time \textbf{P}rompt \textbf{I}ntervention. PI provides an interface to dynamically guide and regulate reasoning paths during inference through timely (\textit{When} module) and proper (\textit{How} module) interventions and post-intervention sampling (\textit{Which} module). This allows human problem-solving expertise and cognitive science principles to be seamlessly integrated into LLMs’ reasoning processes, enhancing controllability and interpretability. Extensive experiments across multiple models and datasets demonstrate that PI significantly shortens CoTs while reducing hallucination, yielding more concise and reliable reasoning.


\end{abstract}


\section{Introduction}

While data and parameter scaling laws \citep{kaplan2020scalinglawsneurallanguage} have long driven progress in large language models (LLMs), the emergence of test-time scaling \citep{snell2024scalingllmtesttimecompute} has shifted the community’s focus toward more efficient methods for enhancing reasoning capabilities. Pioneering works such as OpenAI o1 \citep{o1}, DeepSeek-R1 \citep{deepseekai2025deepseekr1incentivizingreasoningcapability}, and Qwen3 \citep{yang2025qwen3}  exemplify this paradigm, where large reasoning models (LRMs) \citep{xu2025largereasoningmodelssurvey} leverage extended chains of thought (CoTs) \citep{wei2023chainofthoughtpromptingelicitsreasoning}  to tackle complex problems, including mathematics \citep{cobbe2021trainingverifierssolvemathgsm8k, AMC2023, math500hendrycks2021measuringmathematicalproblemsolving, he2024olympiadbenchchallengingbenchmarkpromoting} and scientific \citep{rein2023gpqagraduatelevelgoogleproofqa,minervalewkowycz2022solvingquantitativereasoningproblems} tasks. 

\begin{figure}[!t]
  \centerline{\includegraphics[scale=0.3]{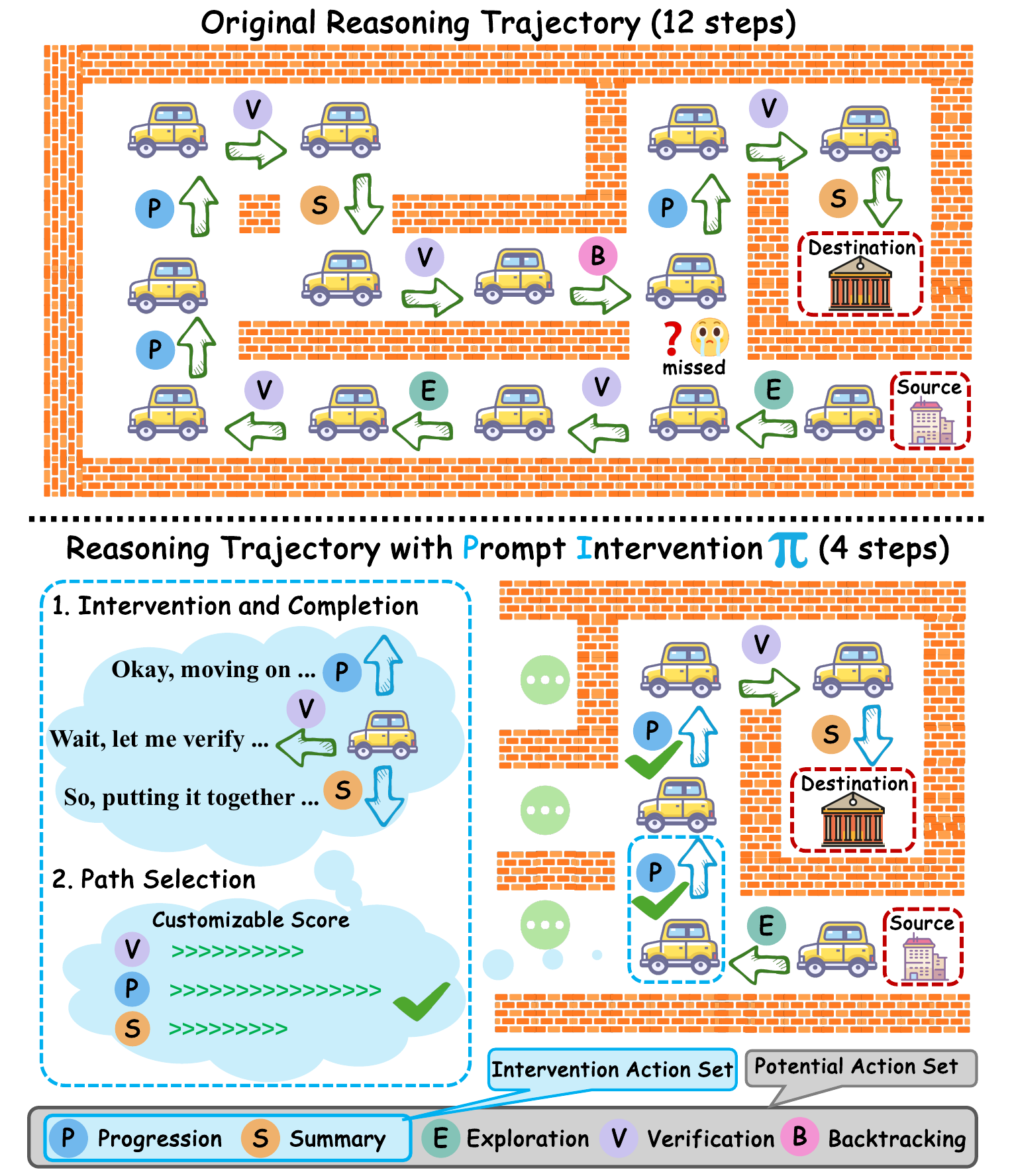}}
  \caption{LRMs' original reasoning misses the optimal trajectory due to overthinking, resulting in verbosity, while $\pi$ 's timely interventions streamline reasoning process, reaching the correct conclusion more efficiently (12 steps to 4 steps).}
  \label{figure1}
\end{figure} 

However, recent studies \citep{chen2025think23overthinkingo1like,team2025kimi}  have revealed that the CoTs generated by these LRMs often exhibit significant redundancy  \citep{chen2025think23overthinkingo1like, cuadron2025danger}, such as repetitive verification and frequent shifts in reasoning \citep{wu2025more, cuadron2025danger,yang2025dynamicearlyexitreasoning}. Some works \citep{manuvinakurike2025thoughtsthinkingreconsideringexplanatory} even argue that ``Chain-of-Thought is not explainability," suggesting that CoTs tend to produce plausible yet non-causal explanations, and advocate for integrating cognitive science principles to make AI explanations more aligned with human reasoning. The root cause lies in the current post-training paradigm of LRMs, which primarily relies on 0/1 outcome rewards, like GRPO \citep{shao2024deepseekmathpushinglimitsmathematical,schulman2017proximalpolicyoptimizationalgorithms}, rather than process-based rewards \citep{lightman2023letsverifystepstep,wang2024mathshepherdverifyreinforcellms,zhang2025lessonsdevelopingprocessreward}, as the latter is inherently difficult to scale for data construction. This results in a lack of regulation over intermediate reasoning steps during training. Intuitively, if empowering models to regulate intermediate reasoning during training is challenging, can we instead intervene in the reasoning process at test time to achieve more concise and reliable CoT?


\begin{figure*}[h!]
  \centerline{\includegraphics[scale=0.38]{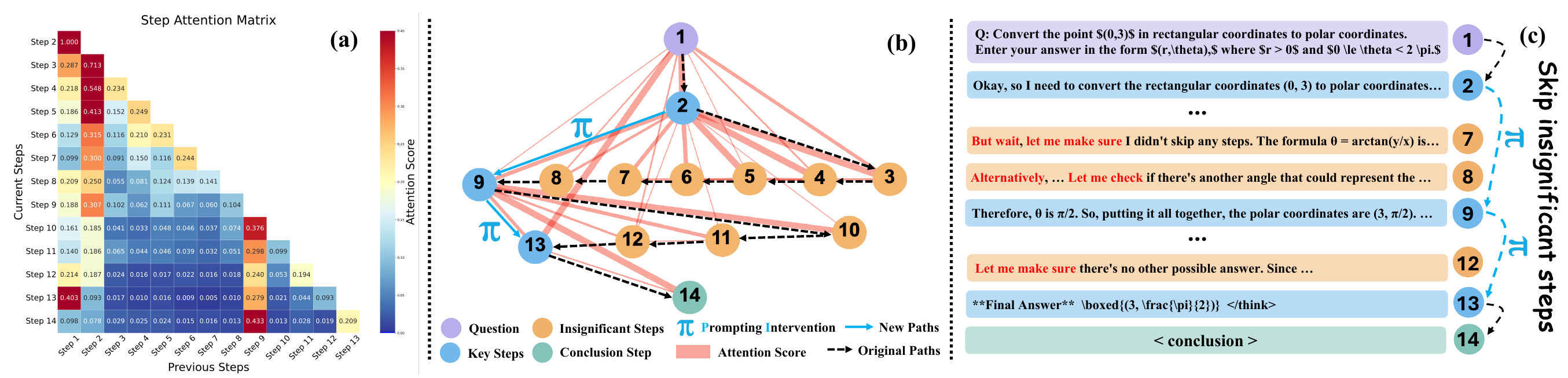}}
  \caption{An illustrative example showing that LRMs exhibit confused and redundant thought processes during reasoning. (a) Step-level attention map visualization of a complete reasoning trajectory, with steps separated by ``\textit{\textbackslash n\textbackslash n}". (b) Directed Acyclic Graph (DAG) representation of the reasoning process, where each step is a node and edge thickness reflects the magnitude of attention values. (c) Abbreviated content of each reasoning step. More details of experimental setups are placed in Appendix A.}
  \label{figure2}
\end{figure*}

As illustrated in Figure 1 (upper), LRMs' original reasoning deviates from the optimal trajectory due to overthinking, resulting in verbosity. However, by intervening at critical moments (e.g., step 2) to guide reasoning actions, as shown in the Figure 1 (lower), we can enforce progressive reasoning and eliminate unnecessary shifts, thereby reaching the correct conclusion more efficiently (reducing steps from 12 to 4). 
While prior work has preliminarily explored inserting prompts during CoT generation (e.g., s1’s \citep{muennighoff2025s1simpletesttimescaling} forced thinking or DEER’s \citep{yang2025dynamicearlyexitreasoning} early exit), none have systematically addressed the guided regulation of reasoning paths. This paper uniquely enables dynamic control over CoT trajectory unfolding, significantly enhancing both the controllability and interpretability \citep{manuvinakurike2025thoughtsthinkingreconsideringexplanatory} in model reasoning.


To this end, we propose a framework, PI ($\pi$), of test-time \textbf{P}rompt \textbf{I}ntervention, which is designed to compensate for the lack of intermediate reasoning regulation during model training. Within this framework, we explicitly define the purpose and function of each reasoning step, (such as verification, summarization, and progressive reasoning) thereby enhancing the transparency and interpretability of the CoT. Building on this foundation, the proposed framework provides an interface for users to dynamically control the model's reasoning path at appropriate junctures. This enables the integration of human problem-solving expertise and cognitive science principles into the CoT generation process, resulting in more concise and reliable CoT. 

Specifically, the PI framework consists of three core modules: \textit{When} Module determines the optimal intervention timing. \textit{How} Module establishes the most effective polices for intervention. \textit{Which} Module chooses the best candidate reasoning path post-intervention. We conducted a systematic analysis of different implementations for these modules, empirically validating the effectiveness of carefully designed fixed-intervention patterns. Furthermore, we propose an automated intervention strategy adaptable to diverse scenarios. Extensive experiments across multiple models and datasets demonstrate that our method significantly improves reasoning conciseness, while effectively mitigating LLM hallucinations \cite{Huang_2025} (see Figure \ref{case-study}). These results underscore the high potential of test-time prompt intervention paradigms.



Our key contributions are summarized as follows:

\begin{itemize}
    \item We propose a novel test-time prompt intervention framework that regulates reasoning processes, offering new perspectives for controllable chain-of-thought generation. 
    \item Our plug-and-play method demonstrates remarkable scalability across models of varying scales (e.g., DeepSeek Series, Qwen3), achieving comparable or superior accuracy with only 40.5\% to 50.4\% of original CoT length on STEM benchmarks (GSM8K, Math500, AMC, OlympiadBench, GPQA, Minerva). Extensive experiments on GSM-NoOp and TruthfulQA datasets show 2.5\%-4.1\% reduction in hallucinations.
    \item The proposed PI framework introduces a human-AI collaboration interface, seamlessly integrating cognitive science principles and expert knowledge to guide LLMs toward more efficient and reliable reasoning paradigms.
\end{itemize}



\section{Observations and Motivations}

In this section, we analyze the reasoning patterns of LRMs and identify issues inherent in their CoTs by visualization and statistical analysis. We then conduct a preliminary exploration of prompt intervention, revealing substantial opportunities for optimization in LRMs' reasoning trajectories.



\begin{figure*}[h!]
  \centerline{\includegraphics[scale=0.39]{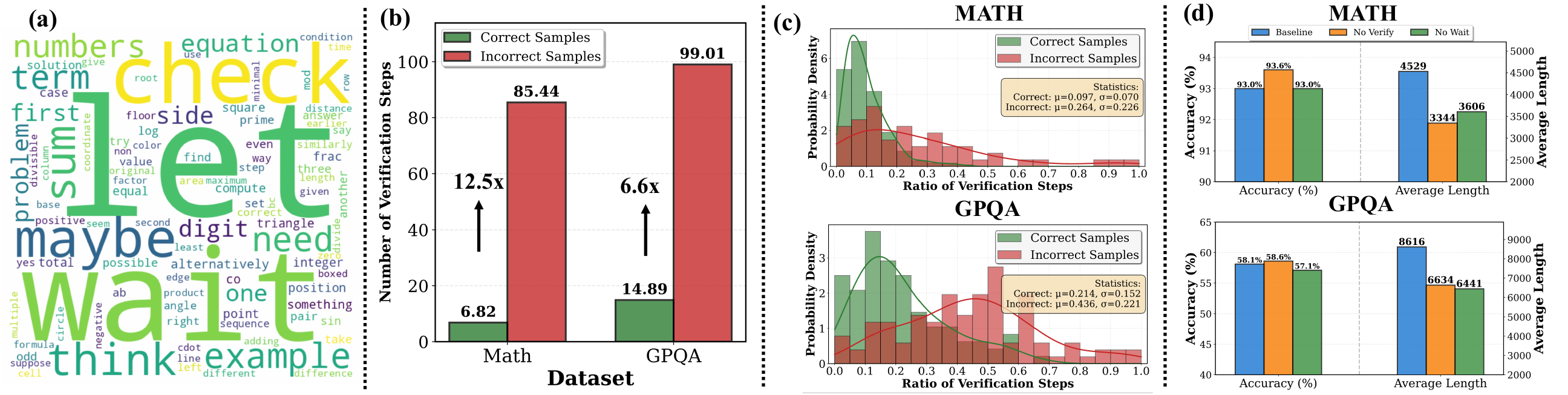}}
  \caption{(a) The word cloud visualization of the LRMs' CoTs. (b) The number of verification steps for correctly answered versus incorrectly answered samples.  (c) The distribution of the proportion of verification steps in correct answers versus incorrect answers. (d) Accuracy and length of Qwen3-8B-generated response on two datasets under different processing strategies. }
  \label{pilot-PI}
\end{figure*} 



While LRMs exploit the test-time scaling law to achieve expert-level performance on complex tasks, recent studies have revealed that they may consume thousands of tokens to solve trivial problems like ``$2+3=?$" \cite{chen2025think23overthinkingo1like} and exhibit problematic reasoning patterns, such as generating more than twice the length for incorrect answers compared to correct ones \cite{fatemi2025concisereasoningreinforcementlearning}. 
This inevitably raises questions about the rationality of LRMs' reasoning trajectories.
To understand LRMs' reasoning behavior, we first seek to answer the question: \textbf{What is the logical structure of CoT when expanded into reasoning steps?}

To investigate this question, we visualize attention maps to reveal how reasoning steps interconnect, providing an intuitive view of dependency structures in the CoTs. We analyze the reasoning trajectory generated by Qwen3-8B on a MATH-500 sample, shown in Figure 2(c). The reasoning process is segmented into steps using ``\textit{\textbackslash n\textbackslash n}" delimiters, with step-level attention scores displayed in Figure 2(a). Based on these attention patterns, we construct a reasoning graph (Figure 2(b)) where nodes connect when their attention score exceeds 0.1, and edge thickness indicates relationship strength. More setups are placed in Appendix A.

Examining Figure 2 collectively, we observe distinct attention patterns throughout the reasoning process. Early stages focus primarily on step 2, which explores the problem-solving approach, while backtracking and verification steps (steps 7-8) receive minimal subsequent attention. After generating step 9 with the correct answer, all following steps predominantly attend to this pivotal moment. However, the model performs several redundant checks with low attention scores (e.g., step 12) before reaching the final conclusion.
We consider steps receiving negligible attention during subsequent reasoning as redundant. Bypassing these through generation intervention could substantially enhance efficiency. Using the graph structure in Figure 2(b), we formalize this analysis by identifying critical steps: a subset where each node includes all its highly-attended predecessors. If the model generated only these critical steps (2, 9, and 13), as shown in Figure 2(c), it would achieve a 75\% reduction in computational overhead.



After visually exploring the logical structure of CoT, we proceed to examine the question: \textbf{What problematic reasoning behaviors in LRMs require intervention?} 


Figure 2(c) shows that low-attention steps are frequently associated with verification processes. Word frequency analysis of LRMs' CoT sequences confirms this pattern—the word clouds in Figure 3(a) reveal high frequencies of verification-related terms like \textit{``wait"} and \textit{``check"}.
We further compared verification step occurrence in correct versus incorrect answers from Qwen3-8B on MATH-500 and GPQA datasets. Figure 3(b) shows the model generates significantly more verification steps for incorrect samples—\textbf{12.5} times more on MATH-500 and \textbf{6.6} times more on GPQA. The histograms in Figure 3(c) illustrate verification step distributions relative to total reasoning steps, revealing that the model achieves higher accuracy on samples with fewer verification steps.
These findings suggest that excessive verification hinders reasoning efficiency, making it harder for LRMs to reach answers. This indicates verification may be the problematic behavior requiring intervention.

\begin{figure*}[h!]
  \centerline{\includegraphics[scale=0.31]{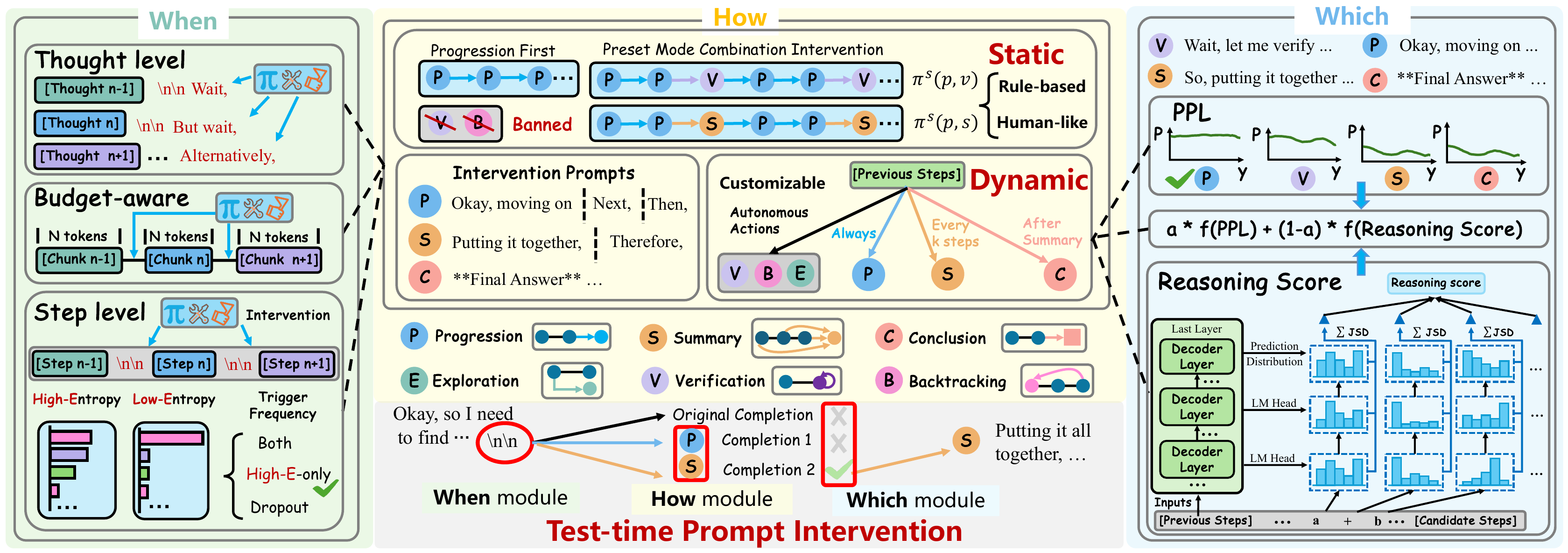}}
  \caption{An overview of the Prompt Intervention ($\pi$) framework. See cases in Appendix B for detailed illustration.}
  \label{figure4}
\end{figure*} 



Building on these findings, we conducted preliminary experiments with two verification intervention strategies and compared their effects on accuracy and length, as shown in Figure 3(d). \texttt{NoVerify} uses post-processing to mask verification steps in generated CoTs, while \texttt{NoWait} employs test-time intervention by replacing verification trigger words during generation. Additional details are provided in Appendix A.
The results of \texttt{NoVerify} demonstrates that removing verification steps from the generated reasoning does not hinder the model's ability to reach correct conclusions. Instead, it helps the model summarize key reasoning information more accurately, achieving higher accuracy with reduced token costs.
The results of \texttt{NoWait} shows that even a simple intervention effectively reduce token consumption while maintaining stable or only slightly reduced accuracy. These outcomes further confirm the existence of redundant verification steps in LRMs' CoTs and highlight the potential for optimization through prompt intervention.

\section{Methodology}


Motivated by the pilot observations, we argue that while LRMs are intelligent thinkers, they frequently exhibit convoluted reasoning patterns due to insufficient meta-cognitive learning during post-training \cite{wang2024metacognitivepromptingimprovesunderstanding, griot2025large, didolkar2024metacognitivecapabilitiesllmsexploration}. This necessitates human guidance as a meta-thinker to appropriately steer the model's thinking processes.
To address this challenge, we introduce the test-time Prompt Intervention ($\pi$) framework (Figure 4), designed to regulate LLM reasoning behavior through strategic human intervention. The framework comprises three interconnected modules: determining when to intervene (\textit{\textbf{When}} module), establishing how to intervene (\textit{\textbf{How}} module), and selecting which post-intervention reasoning path to adopt (\textit{\textbf{Which}} module).



\subsection{\textit{How} Module}
\paragraph{Reasoning Behaviors.} To steer the CoT in LRMs more scientifically, we first analyze their reasoning behaviors.
Inspired by recent work \cite{gandhi2025cognitivebehaviorsenableselfimproving, luo2025deconstructinglongchainofthoughtstructured} and based on observations of the generated CoTs, we categorize reasoning steps into six types: \textbf{Progression}, \textbf{Summary}, \textbf{Exploration}, \textbf{Verification}, \textbf{Backtracking}, and \textbf{Conclusion}.
\begin{itemize}
\item \textbf{Progression} involves advancing further along the current line of reasoning based on known information and inference rules, often accompanied by connective words such as ``\textit{Next}", ``\textit{Then}" or phrases like ``\textit{Okay, moving on}".

\item \textbf{Summary} involves organizing and integrating key information obtained from existing reasoning steps to lay the foundation for subsequent reasoning, often accompanied by summarizing phrases such as ``\textit{Putting it together}".

\item \textbf{Exploration} involves actively generating new hypotheses or seeking alternative solution approaches when the current reasoning trajectory fails to yield progress, often accompanied by connective words like ``\textit{Alternatively}".

\item \textbf{Verification} involves checking and confirming the logical consistency and accuracy of recently generated reasoning steps, typically accompanied by ``\textit{Wait}".

\item \textbf{Backtracking} enables the system to revert to earlier decision points and select new paths when the current reasoning approach is incorrect, facilitating error correction.

\item \textbf{Conclusion} delivers the final answer once adequate and accurate reasoning information has been gathered.
\end{itemize}


Since these behaviors in LLMs are frequently accompanied by action-triggering signals, we can strategically insert different trigger words during the reasoning process to intervene in the model's reasoning trajectory. Based on the intervention mechanisms employed, PI ($\pi$) can be categorized into static intervention and dynamic intervention approaches. Static interventions are well-suited for deliberate design incorporating cognitive theoretical frameworks, while dynamic interventions demonstrate superior generalizability and broader applicability across diverse tasks.








\begin{figure}[!t]
  \centerline{\includegraphics[scale=0.48]{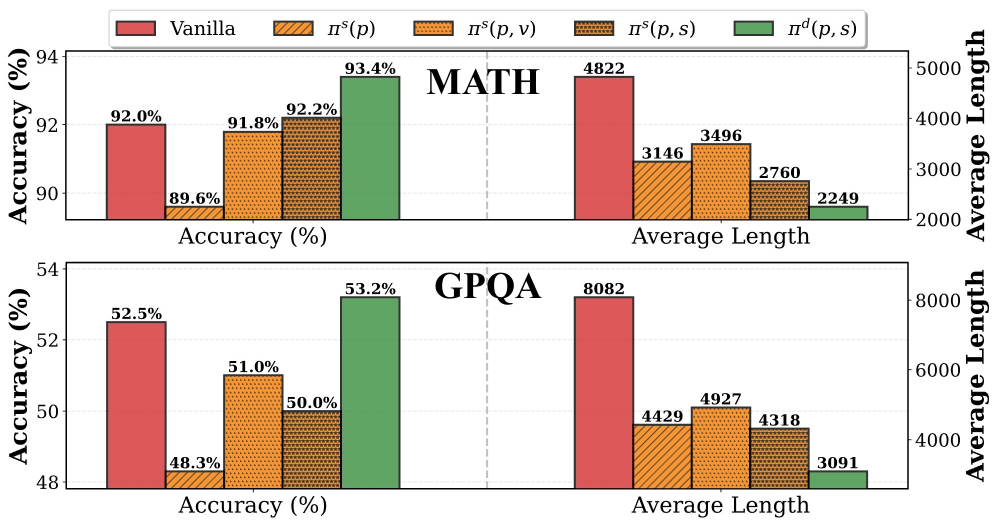}}
  \caption{Comparison of experimental results on Qwen3-4B between original generation, static PI, and dynamic PI.}
  \label{figure5}
\end{figure} 

\paragraph{Static Intervention.} S1 \cite{muennighoff2025s1simpletesttimescaling} represents a special case of static intervention, which incorporates additional verification and exploration steps. To address the overthinking problem, developed several static PI strategies as shown in Figure 4 to reduce verification.
Figure 5 shows the performance of multiple predefined static intervention strategies, including progressive priority ($\pi^s(p)$), progressive with verification ($\pi^s(p,v)$), progressive with summarization ($\pi^s(p,s)$). Experimental results demonstrate length declines on simple problems, whereas accuracy drops on challenging questions. This suggests that while static PI mitigates overthinking issues in simple cases, the rigid predefined intervention patterns hinder the model's reasoning ability when dealing with complex problems. 


\paragraph{Dynamic Intervention.}  Given the substantial variability across problems, it becomes challenging to predetermine the optimal reasoning trajectory for each specific instance. To address these limitations, we develop dynamic PI strategies that mitigate the risk of over-intervention. Specifically, upon completion of a reasoning step, dynamic PI concurrently extends multiple branches that generate diverse reasoning behaviors. These are combined with the model's naturally generated reasoning steps as candidate options, with the optimal path selected using the \textit{Which} module design.
\begin{equation}
  \mathbf{S}^{t+1}=\{\mathbf{S}_i^{t+1}\}, \   \mathbf{S}_i^{t+1} = \text{LRM}(\mathbf{S}^{\leq t},\mathbf{T}_i), \ \mathbf{T}_i \in \mathcal{T},
\end{equation}
where $S_i$ is the candidate step and $\mathcal{T}$ denotes trigger sets.
A key advantage of dynamic PI lies in its ability to flexibly adapt intervention actions based on varying task demands. When prioritizing reasoning efficiency, we designate progression behavior as a constant candidate action, invoke summary behavior less frequently, and preserve other reasoning behaviors that emerge naturally from the model, thus promoting depth-first reasoning in CoT ($\pi^d(p,s)$). For simple tasks, conclusion behavior can be added to facilitate early exit, further mitigating overthinking ($\pi^d(p,s,c)$). For trust-critical applications, verification branch can be incorporated to reduce hallucinations ($\pi^d(p,s,v)$). Once dynamic PI generates multiple branches,  the choice of optimal branch (determined by the \textit{Which} module) and intervention timing (governed by the \textit{When} module) becomes crucial.



\subsection{\textit{Which} Module}






A straightforward approach involves relying on LRM's prediction confidence by evaluating the perplexity (PPL) of candidates and selecting the branch with the lowest PPL.
\begin{equation}
    \text{PPL}(\mathbf{S}_i) = \exp(-\frac{1}{|\mathbf{S}_i|}\sum_{y_{t} \in \mathbf{S}_i} \log P(y_{t})).
\end{equation}


Nevertheless, branch selection based purely on perplexity can lead the model into degenerative behaviors such as repetitive patterns.
To address this limitation, we seek a metric that captures ``reasoning depth" to guide branch selection. By prioritizing branches with deeper reasoning, the \textit{Which} module minimizes superficial information propagation and accelerates the reasoning process. We characterize token-level decision disparities through differences in probability distributions between the model's early layers and the final layer, where large disparities indicate critical nonlinear integration and reasoning occurring across layers. By conceptualizing the hidden state space of multi-layer Transformers as a high-dimensional semantic manifold, we provide theoretical analysis from geometric and causal perspectives, accompanied by visual illustrations in Appendix B. In practice, we quantify this disparity using Jensen-Shannon divergence (JSD) \cite{chuang2024doladecodingcontrastinglayers,sun2025detectionmitigationhallucinationlarge}, computing the Reasoning Depth Score (RDS) by averaging across all tokens over multiple early layers in each step, as follows: 
\begin{equation}
R(\mathbf{S}_i) = \frac{1}{|\mathbf{S}_i|} \sum_{y_{t} \in \mathbf{S}_i} \frac{1}{|\mathcal{L}|} \sum_{l \in \mathcal{L}} \operatorname{JSD}(p(y_{t}), q_{j}(y_{t})),
\label{eq1}
\end{equation}
\begin{equation}
q_{l}(y_{t}) = \operatorname{softmax}\left(\operatorname{LM-Head}\left(h_{t}^{(l)}\right)\right), \quad l \in \mathcal{L},
\label{eq2}
\end{equation}
where $\mathcal{L}$ is the set of selected early layers and $p(y_{t})$ is the final distribution from the last layer. 

After separately normalizing both perplexity and RDS, we select the optimal reasoning branch based on their combined score $S = \arg\max_{i=1\dots k} P(S_i)$, calculated as:
\begin{equation}
    P(\mathbf{S}_i) = \alpha \cdot \text{Norm}(\text{PPL}(\mathbf{S}_i)^{-1}) + \beta \cdot \text{Norm}( R(\mathbf{S}_i)),
\end{equation}
where Norm denotes normalization and $\beta=1-\alpha$.
This scoring design considers both the logical coherence and thinking depth of reasoning steps. Based on a greedy strategy, it strengthens the model's reasoning at each step, refraining superficial information transfer, thus arriving at the final conclusion more efficiently.

{
\setlength{\tabcolsep}{2.5pt}
\begin{table*}[h!]
\centering
\scalebox{0.82}{
\begin{tabular}{@{}lcccccccccccccccccccc@{}} 
\toprule
 \multirow{2}{*}{\textbf{Method}} 
 & \multicolumn{3}{|c}{\textbf{GSM8K}}& \multicolumn{3}{c}{\textbf{MATH-500}} & \multicolumn{3}{c}{\textbf{AMC}} & \multicolumn{3}{c}{\textbf{OlympiadBench}}  & \multicolumn{3}{c}{\textbf{GPQA-D}}  & \multicolumn{3}{c}{\textbf{Minerva}} & \multicolumn{2}{|c}{\textbf{Overall}} \\
   & {Acc$\uparrow$} & {Tok$\downarrow$} & {CR$\downarrow$} & {Acc$\uparrow$} & {Tok$\downarrow$} & {CR$\downarrow$} & Acc$\uparrow$ & Tok$\downarrow$ & {CR$\downarrow$} & {Acc}$\uparrow$  & {Tok$\downarrow$} & {CR$\downarrow$} & {Acc$\uparrow$} & {Tok$\downarrow$} & {CR$\downarrow$} & {Acc$\uparrow$} & {Tok$\downarrow$} & {CR$\downarrow$}& {Acc$\uparrow$} & {CR}$\downarrow$  \\ 
\hline

\multicolumn{21}{l}{{\cellcolor[rgb]{0.957,0.957,0.957}}\textit{\textbf{DeepSeek-R1-Distill-Qwen-14B}}} \\
\textit{Vanilla} & 93.9 & 1,458 & 100\% & 90.0 & 4,012 &100\% & 87.5 & 6,958 & 100\% & 55.6 & 8,063 & 100\% & 53.9 & 7,132 & 100\% & 45.6 & 4,932 & 100\%  & \multicolumn{1}{|l}{71.1} & 100\%  \\
\textit{NoThinking} & 90.1 & 272 & 18.7\% & 76.2 & 646 & 16.1\% & 65.0 & 1,106  & 15.9\% & 42.7 & 1,711 & 21.2\% & 38.4 & 548 & 7.8\% & 38.3 & 549 & 11.1\%   & \multicolumn{1}{|l}{58.5} & \textbf{15.1\%}  \\
\textit{NOWAIT} & 92.4 & 503 & 34.5\% & 88.2 & 2,524 & 62.9\% & 90.0 & 4,320  & 62.1\% & 56.1 & 5,913 & 73.3\% & 51.0 & 4,227 & 59.3\% & 44.2 &  2,865 & 58.1\%   & \multicolumn{1}{|l}{70.3} & 58.4\%  \\
\textit{DEER} & 93.3 & 1,006 & 69.0\% & 90.2 & 2,457 & 61.2\% & 88.8 & 4,196 & 60.3\% & 55.0 & 5,695  & 70.6\% & 56.3 & 4,628 & 64.9\% & 46.0 & 3,154 & 64.0\%   & \multicolumn{1}{|l}{\textbf{71.6}} & 65.0\%  \\
\rowcolor[rgb]{0.87,0.94,1}
\textit{PI-$\pi^d(p,s)$} & 93.9 & 572 & 39.2\% & 89.6 & 2,042 & 50.9\% & 92.5 & 3,736 & 53.7\% & 58.7 & 5,098 & 63.2\% & 55.8 & 3,906 & 54.8\% & 46.7 & 1,997 & 40.5\%  & \multicolumn{1}{|l}{\textbf{72.9}} & \textbf{50.4\%}  \\
\hline

\multicolumn{21}{l}{{\cellcolor[rgb]{0.957,0.957,0.957}}\textit{\textbf{Qwen3-4B}}} \\
\textit{Vanilla} & 94.8 & 2,156 & 100\% & 92.0 & 4,822 &100\% & 87.5 & 8,002 & 100\% & 59.7 & 9,128 & 100\% & 52.5 & 8,082 & 100\% & 50.0 & 6,583 & 100\%  & \multicolumn{1}{|l}{72.8} & 100\%  \\
\textit{NoThinking} & 92.5 & 320 & 14.8\% & 85.6 & 1,079 & 22.4\% & 68.8 & 2,413  & 30.2\% & 52.6 & 2,571 & 28.2\% & 43.4 & 1,507 & 18.6\% & 43.0 & 747 & 11.3\%   & \multicolumn{1}{|l}{64.3} & \textbf{20.9\%}  \\
\textit{NOWAIT} & 94.6 & 1,176 & 54.5\% & 93.0 & 3,893 & 80.7\% & 87.5 & 7,181  & 89.7\% & 61.0 & 8,105 & 88.8\% & 52.9 & 7,232 & 89.5\% & 50.7 &  5,401 & 82.0\%   & \multicolumn{1}{|l}{73.3} & 80.9\%  \\
\textit{DEER} & 94.9 & 1,088 & 50.5\% & 93.0 & 3,424 & 71.0\% & 87.5 & 4,906 & 61.3\% & 64.6 & 7,454  & 81.7\% & 53.6 & 7,353 & 91.0\% & 50.5 & 4,001 & 60.8\%   & \multicolumn{1}{|l}{\textbf{74.0}} & 69.4\%  \\
\rowcolor[rgb]{0.87,0.94,1}
\textit{PI-$\pi^d(p,s)$} & 95.2 & 596 & 27.6\% & 93.4 & 2,249 & 46.6\% & 90.0 & 3,740 & 46.7\% & 64.2 & 4,711 & 51.6\% & 53.2 & 3,091 & 38.2\% & 50.0 & 2,118 & 32.2\%  & \multicolumn{1}{|l}{\textbf{74.3}} & \textbf{40.5\%}  \\
\hline

\multicolumn{21}{l}{{\cellcolor[rgb]{0.957,0.957,0.957}}\textit{\textbf{Qwen3-8B}}} \\

\textit{Vanilla} & 95.2 & 2,191 & 100\% & 92.4 & 5,224 & 100\% & 88.8 & 8,027 & 100\% & 60.3 & 9,414 & 100\% & 58.1 & 9,105 & 100\% & 52.4 & 6,850 & 100\%  & \multicolumn{1}{|l}{74.5} & 100\%  \\
\textit{NoThinking} & 93.3 & 304 & 13.9\% & 85.8 & 1,052 & 20.1\% & 72.5 & 2,397  & 29.9\% & 51.0 & 2,503 & 26.6\% & 52.0 & 1,551 & 17.0\% & 44.5 & 664 & 9.7\%   & \multicolumn{1}{|l}{66.5} & \textbf{19.5\%}  \\
\textit{NOWAIT} & 95.0 & 1,220 & 55.7\% & 93.2 & 4,007 & 76.7\% & 87.5 & 7,181 & 89.5\% & 63.0 & 8,294  & 88.1\% & 57.3 & 7,722 & 84.8\% & 51.8 & 5,658 & 82.6\%   & \multicolumn{1}{|l}{74.6} & 79.6\%  \\
\textit{DEER} & 95.5 & 1,042 & 47.6\% & 92.2 & 3,124 & 59.8\% & 88.8& 4,486 & 55.9\% & 65.0 & 7,357  & 78.1\% & 59.1 & 8,596 & 94.4\% & 52.2 & 3,941 & 57.5\%   & \multicolumn{1}{|l}{\textbf{75.5}} & 65.6\%  \\
\rowcolor[rgb]{0.87,0.94,1}
\textit{PI-$\pi^d(p,s)$} & 95.3 & 840 & 38.8\% & 94.0 & 3,074 & 58.8\% & 89.4 & 4,814 & 60.0\% & 65.5 & 5,573 & 59.2\% & 58.6 & 4,081 & 44.8\% & 52.0 & 2,724 & 39.8\% & \multicolumn{1}{|l}{\textbf{75.8}} & \textbf{50.2\%}  \\
\hline



\multicolumn{21}{l}{{\cellcolor[rgb]{0.957,0.957,0.957}}\textit{\textbf{Qwen3-14B}}} \\
\textit{Vanilla} & 95.8 & 1,642 & 100\% & 94.2 & 4,540 & 100\% & 93.8 & 6,755 & 100\% & 64.3 & 8,778 & 100\% & 60.1 & 7,694 & 100\% & 54.2 & 5,776 & 100\%  & \multicolumn{1}{|l}{{77.1}} & 100\%  \\
\textit{NoThinking} & 95.1 & 278 & 16.9\% & 88.2 & 852 & 18.8\% & 77.5 & 2,065  & 30.6\% & 51.7 & 2,029 & 23.1\% & 55.6 & 1,286 & 16.7\%  & 43.4 & 649 & 11.2\% & \multicolumn{1}{|l}{68.6} & \textbf{19.6\%}  \\
\textit{NOWAIT} & 96.1 & 1,011 & 61.6\% & 94.0 & 3,783 & 83.3\% & 94.4 & 6,855  & 101.5\% & 64.0 & 7,939 & 90.4\% & 59.4 & 6,635 & 86.2\%  & 55.9 & 4,830 & 83.6\% & \multicolumn{1}{|l}{77.3} & 84.4\%  \\
\textit{DEER} & 95.8 & 929 & 56.6\% & 94.2 & 2,953 & 65.0\% & 95.0 & 4,813 & 71.3\% & 65.1 & 6,777 & 77.2\% & 60.3 & 7,261 & 94.4\% & 54.0 & 3,552 & 61.5\% & \multicolumn{1}{|l}{\textbf{77.4}} & 
{70.1\%}  \\
\rowcolor[rgb]{0.87,0.94,1}
\textit{PI-$\pi^d(p,s)$} & 96.0 & 542 & 33.0\% & 95.0 & 2,515 & 55.4\% & 95.0 & 4,146 & 61.4\% & 66.5 & 5,128 & 58.4\% & 59.6 & 3,320 & 43.2\% & 53.3 & 1,907 & 33.0\% & \multicolumn{1}{|l}{\textbf{77.6}} & \textbf{47.4\%}  \\

 \bottomrule
\end{tabular}
}
\caption{Experimental results on various LRMs. "Acc" denotes accuracy, "Tok" denotes token count, and "CR" denotes compression rate. $\uparrow$/$\downarrow$ indicate that higher/lower values are better. The top-2 best results are highlighted in \textbf{bold}. The result is statistically significant with $p$-value $<$ 0.05.}
\label{baselines}
\end{table*}
}
\subsection{\textit{When} Module}

As for intervention timing, it can be configured at various granularities, such as fixed token intervals or at natural transition points in the reasoning process \citet{yang2025dynamicearlyexitreasoning}.
In this work, PI adopts step-level intervention, using ``\textit{\textbackslash n\textbackslash n}" as the step delimiter. However, relying solely on the explicit ``\textit{\textbackslash n\textbackslash n}" for segmentation and expanding branch at every step presents limitations. These limitations arise from two key factors: first, the inherent uncertainty in step granularity, as a single major step may encompass multiple sub-steps; and second, the potential strong correlations between adjacent steps, where subsequent steps often represent logical consequences of their predecessors. Inspired by \citet{wang20258020rulehighentropyminority}, we combine the model's internal state, specifically entropy, to determine optimal intervention timing.
\begin{equation}
    H(y_0) = -\sum_{y_0\in\mathcal{V}} p\left(y_0\right) \log p\left(y_0\right),
\end{equation}
where $y_0$ denotes the first token that LRMs generate at the current step. We provide theoretical analysis and illustrative examples demonstrating the performance advantages (see Appendix B) and efficiency improvements (see Appendix F) of entropy-based step-level intervention.


%








\section{Experiments}

\newcommand{\annotate}[3]{%
    #1\raisebox{-0.5ex}{\scriptsize\textcolor{#2}{#3}}%
}

{
\setlength{\tabcolsep}{2.5pt}
\begin{table}[!t]
\centering
\scalebox{0.9}{
\begin{tabular}{@{}lcccccccccccccccccccc@{}} 
\toprule
 \multirow{2}{*}{\textbf{Method}} 
 & \multicolumn{2}{|c}{\textbf{GSM-NoOp}}& \multicolumn{4}{|c}{\textbf{TruthfulQA}}  \\
   & {Acc$\uparrow$} & {Tokens$\downarrow$}  & {MC1$\uparrow$} & {Tokens$\downarrow$} & {MC2$\uparrow$} & {Tokens$\downarrow$} \\ 
\hline

\textit{Vanilla} & 82.5 & 1998 & 56.0 & 673 & 70.2 & 766  \\
\textit{NoThinking} & 77.1 & 317 & 14.8 & 147 & 2.9 & 184 \\
\textit{NOWAIT} & 79.3 & 626 & 46.9 & 709 & 60.1 & 774 \\
\textit{DEER} & 83.3 & 1282 & 2.9 & 667 & 5.3 & 810 \\
\textit{PI-$\pi^d(p,s)$} & \textbf{85.0} & 677 & 58.3 & 511 & 72.1 & 543 \\
\textit{PI-$\pi^d(p,s,v)$} & 83.7 & 802 & \textbf{59.1} & 672 & \textbf{74.3} & 739 \\

 \bottomrule
\end{tabular}
}
\caption{Experimental results of PI on hallucination benchmarks with DeepSeek-R1-Distill-Qwen-14B.}
\label{baselines-hallu}
\end{table}
}

\subsection{Experimental Setup}
\paragraph{Benchmarks and Metrics.}
We evaluate model performance across eight benchmarks spanning three categories: four mathematical reasoning benchmarks (GSM8K \citep{cobbe2021trainingverifierssolvemathgsm8k}, MATH-500 \citep{math500hendrycks2021measuringmathematicalproblemsolving}, AMC 2023 \citep{AMC2023}, and OlympiadBench \citep{he2024olympiadbenchchallengingbenchmarkpromoting}), two STEM reasoning benchmarks (GPQA Diamond \citep{rein2023gpqagraduatelevelgoogleproofqa} and Minerva \citep{minervalewkowycz2022solvingquantitativereasoningproblems}), and two hallucination-related benchmarks (GSM-NoOp \cite{GSMnoopmirzadeh2024gsmsymbolicunderstandinglimitationsmathematical} and TruthfulQA \cite{lin2022truthfulqameasuringmodelsmimic}). 
For evaluation, we employ three metrics: \textit{Accuracy} (\textbf{Acc}), \textit{Token Number} (\textbf{Tok}), and \textit{Compression Rate} (\textbf{CR}). 

\paragraph{Backbone LRMs, Baselines, and Implementations.} We conducted experiments on the Qwen3 series of models (4B, 8B, 14B) \citep{yang2025qwen3technicalreport}, and the DeepSeek-R1-Distill models (including Qwen-7B, Qwen-14B, Llama-8B) \citep{deepseekai2025deepseekr1incentivizingreasoningcapability}. Our model selection spans different sizes and training datasets to validate the robustness and generalizability of PI.
We compare our PI against SoTA training-free efficient reasoning methods, including \textit{NoThinking} \cite{ma2025reasoning}, \textit{NOWAIT} \cite{wang2025waitdontneedwait}, and \textit{DEER} \cite{yang2025dynamicearlyexitreasoning}. 
For the decoding strategy, we employ top-p sampling with the officially recommended parameters of $temperature=0.6$
 and $p=0.95$. We set the maximum generation length to 16384, and set the $\alpha$ to $0.6$. 
More details of benchmarks, metrics, baselines, and implementations are placed in Appendix C. 

\subsection{Experimental Results}
\paragraph{Efficient Reasoning.} Table 1 demonstrates PI's performance across six widely accepted benchmarks on 4 different state-of-the-art reasoning models, demonstrating significant improvements in both accuracy and efficiency. Specifically, compared to vanilla CoT, PI achieves an average accuracy improvement of 0.5 to 1.8 percentage points while reducing sequence length by 49.6\% to 59.6\%. 
Compared to other baselines, PI demonstrates more balanced and comprehensive performance, achieving Pareto-optimal results across both accuracy and compression rate dimensions. We provide additional analysis of computational cost in Appendix F to further demonstrate the efficiency of PI.

\begin{figure*}[h!]
  \centerline{\includegraphics[scale=0.3]{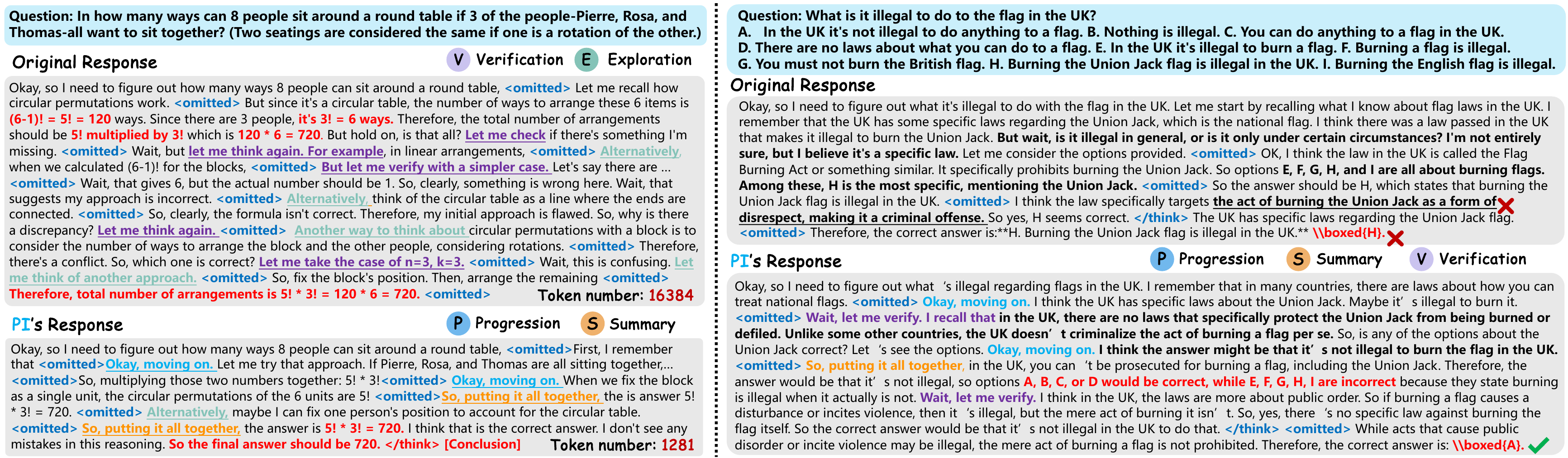}}
  \caption{Comparison of responses generated using PI versus Vanilla CoT on MATH-500 (left) and TruthfulQA (right).}
  \label{case-study}
\end{figure*}

\paragraph{Hallucination.} Table 2 shows that PI effectively reduces hallucination issues by 2.5\% to 4.1\% on two hallucination benchmarks. Specifically, on GSM-NoOp, PI-$\pi^d(p,s)$ mitigates harmful reflection arising from the model's attention to distracting information in problem statements, effectively suppressing overthinking to avoid reasoning hallucinations. On TruthfulQA, a factual dataset that does not involve complex reasoning, PI-$\pi^d(p,s,v)$ improves reliability by incorporating a verification branch that dynamically validates recalled knowledge during the reasoning process.

\begin{figure}[!t]
  \centerline{\includegraphics[scale=0.55]{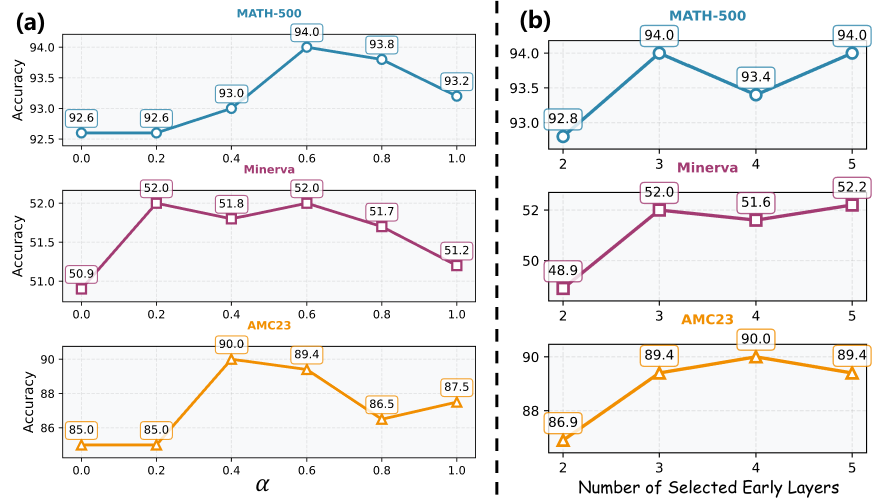}}
  \caption{Performance trends of PI on Qwen3-8B regarding hyperparameter $\alpha$ and number of selected early layers $N$.}
  \label{figure7}
\end{figure} 


\paragraph{Ablation Study.} To validate the effectiveness of the specific designs in PI's three modules (When, Which, and How), we performed ablation studies in Table 3. Specifically, -When(Ent) denotes removing the design that dynamically intervenes only at high-entropy positions, -RDS denotes removing the RDS-based branch selection design. We also tested adding verification \textit{$\pi^d(p,s,v)$} and conclusion \textit{$\pi^d(p,s,c)$} branches. Results show that removing the high-entropy intervention mechanism (-When(Ent)) leads to decreased accuracy, as forcing new branch creation at every step may compromise the reasoning quality of branches. Similarly, excluding reasoning depth score (-RDS) diminishes the model's reasoning depth at each step, leading to reduced accuracy. Adding verification branches proves unnecessary, as the model's spontaneous verifications are sufficient. The conclusion branch variant achieves shorter reasoning while improving performance on GSM8K and AMC, but significantly degrades accuracy on challenging datasets. These findings suggest designing targeted interventions for different scenarios. For instance, retaining the conclusion branch for simple problems while discarding it for challenging problems would maximize PI's benefits.




{
\setlength{\tabcolsep}{2.5pt}
\begin{table*}[h!]
\centering
\scalebox{0.82}{
\begin{tabular}{@{}lcccccccccccccccccccc@{}} 
\toprule
 \multirow{2}{*}{\textbf{Method}} 
 & \multicolumn{3}{|c}{\textbf{GSM8K}}& \multicolumn{3}{c}{\textbf{MATH-500}} & \multicolumn{3}{c}{\textbf{AMC}} & \multicolumn{3}{c}{\textbf{OlympiadBench}}  & \multicolumn{3}{c}{\textbf{GPQA-D}}  & \multicolumn{3}{c}{\textbf{Minerva}} & \multicolumn{2}{|c}{\textbf{Overall}} \\
   & {Acc$\uparrow$} & {Tok$\downarrow$} & {CR$\downarrow$} & {Acc$\uparrow$} & {Tok$\downarrow$} & {CR$\downarrow$} & Acc$\uparrow$ & Tok$\downarrow$ & {CR$\downarrow$} & {Acc}$\uparrow$  & {Tok$\downarrow$} & {CR$\downarrow$} & {Acc$\uparrow$} & {Tok$\downarrow$} & {CR$\downarrow$} & {Acc$\uparrow$} & {Tok$\downarrow$} & {CR$\downarrow$}& {Acc$\uparrow$} & {CR}$\downarrow$  \\ 
\hline

\textit{Vanilla} & 95.2 & 2191 & 100\% & 92.4 & 5224 & 100\% & 88.8 & 8027 & 100\% & 60.3 & 9414 & 100\% & 58.1 & 9105 & 100\% & \textbf{52.4} & 6850 & 100\%  & \multicolumn{1}{|l}{74.5} & 100\%  \\
\textit{$\pi^d(p,s)$} & 95.3 & 840 & 38.8\% & \textbf{94.0} & 3,074 & 58.8\% & 89.4 & 4814 & 60.0\% & \textbf{65.5} & 5573 & 59.2\% & \textbf{58.6} & 4081 & 44.8\% & 52.0 & 2724 & 39.8\% & \multicolumn{1}{|l}{\textbf{75.8}} &{50.2\%}  \\

-\textit{When} (Ent) & 95.2 & 717 & 32.7\% & 93.4 & 2887 & 55.3\% & 90.0 & 4607 & 57.4\% & 64.4 & 5481 & 58.2\% & 57.3 & 4138 & 45.5\% & 52.1 & 2792 & 40.8\% & \multicolumn{1}{|l}{{75.4}} & {48.3\%}  \\

-RDS & \textbf{95.6} & 747 & 34.1\% & 93.2 & 2922 & 55.9\% & 87.5 & 5072 & 63.2\% & 63.7 & 5729 & 60.9\% & 55.3 & 4314 & 47.4\% & 51.2 & 2870 & 41.9\% & \multicolumn{1}{|l}{{74.4}} & {50.6\%}  \\

\textit{$\pi^d(p,s,v)$} & 95.0 & 871 & 39.8\% & 92.6 & 3326 & 63.7\% & 90.0 & 5203 & 64.8\% & 63.4 & 5809 & 61.7\% & 56.5 & 4847 & 53.2\% & 51.9 & 3241 & 47.3\% & \multicolumn{1}{|l}{{74.9}} & {55.1\%}  \\

\textit{$\pi^d(p,s,c)$} & \textbf{95.6} & 696 & 31.8\% & 89.4 & 2054 & 39.3\% & \textbf{91.3} & 3390 & 42.2\% & 55.3 & 3448 & 36.6\% & 46.5 & 1842 & 20.2\% & 47.8 & 1965 & 28.7\% & \multicolumn{1}{|l}{{71.0}} & \textbf{33.0\%}  \\

 \bottomrule
\end{tabular}
}
\caption{Ablation study results on Qwen3-8B. The result is statistically significant with $p$-value $<$ 0.05.}
\label{ablation-all}
\end{table*}
}

\paragraph{Impact of hyperparameters.} Figure 7 illustrates how $\alpha$ and $N$ affect PI performance across multiple benchmarks, revealing that optimal performance is achieved with $\alpha$ ranging from 0.4 to 0.6 and $N$ ranging from 3 to 5. When $\alpha$ values are extremely small (prioritizing reasoning depth only) or large (emphasizing logical coherence only), accuracy suffers, confirming that effective reasoning requires balancing both coherence and reasoning depth. When $N$ exceeds 3, the reasoning depth score (RDS) computation adequately captures thinking depth, facilitating the generation of more concise and effective reasoning chains in LRMs.



\paragraph{Case Study.} The left side of Figure 6 shows the responses of PI and Vanilla CoT on a math problem. Through artificial intervention during generation, PI enhances the priority of progression and summarization while substantially reducing verification behaviors, thus reaching the final answer with reduced reasoning costs. Although Vanilla CoT also obtains the correct answer, it exhibits overthinking with more than ten verification attempts or thought switches, revealing the disorganized thought patterns in LRMs.
The right side of Figure 6 presents a comparison on factual QA. Initially, both methods assume that burning the British flag is illegal. However, through reflection, PI recalls that unlike other countries, the UK does not have specific laws protecting the flag. After summarizing the CoT, PI guides the LRM to the correct answer. Additionally, by integrating with the online Human-AI Collaborative Reasoning Interface mentioned in Appendix G, PI can enhance the interpretability of LRM reasoning steps. More examples are provided in Appendix E.






\section{Related Work}

The original intention of test-time computing \cite{snell2024scalingllmtesttimecompute}  is to enhance the intelligence of LLMs by increasing computational overhead and latency during the inference phase. Following the test-time scaling law \cite{ji2025surveytesttimecomputeintuitive}, two directions have garnered significant attention in the LLM community: test-time training \cite{sun2020testtimetrainingselfsupervisiongeneralization} and test-time reasoning \cite{zhang2024restmctsllmselftrainingprocess, dai2025sgrpoearlyexitreinforcement, dai2025stablereinforcementlearningefficient}. The former enables models to tackle data distributional shifts \cite{liu2021ttt++,zhu2024efficienttesttimeprompttuning}  and learn to memorize \cite{behrouz2024titanslearningmemorizetest} by updating parameters during testing, while the latter improves the model's reasoning capabilities through techniques like repeat Sampling \cite{wang2023selfconsistencyimproveschainthought,gulcehre2023reinforcedselftrainingrestlanguage,gui2024bonbonalignmentlargelanguage,amini2025variationalbestofnalignment}, self-correlation \cite{paul2024refinerreasoningfeedbackintermediate,gou2024criticlargelanguagemodels,welleck2022generatingsequenceslearningselfcorrect,havrilla2024glorewhenwhereimprove} and tree search \cite{yao2023treethoughtsdeliberateproblem,qi2024mutualreasoningmakessmaller,zhang2024restmctsllmselftrainingprocess,hao2023reasoninglanguagemodelplanning}. Some works focus on optimizing demonstrations \cite{chen2023selficlzeroshotincontextlearning,kim2022selfgeneratedincontextlearningleveraging} or feedback modeling \cite{zheng2023judgingllmasajudgemtbenchchatbot,li2023generativejudgeevaluatingalignment,yang2024orthogonalfinetuningdirectpreference} at test time.

This paper pioneers a novel direction in test-time computing: test-time prompt intervention. Unlike existing paradigms, its distinctive feature is providing an interface for human-LLM interaction, enabling human expertise and cognitive science insights \cite{manuvinakurike2025thoughtsthinkingreconsideringexplanatory} to guide the model's CoT generation for specific tasks. This paradigm unlocks the model's full potential across domains while improving controllability and reducing hallucinations.


\section{Conclusion and Future Work}

This paper proposes a new research branch of test-time compute: Test-Time Prompt Intervention (PI), which guides models through interventions during inference to generate more controllable, concise, and less hallucinatory CoTs. The PI framework formalizes and empirically validates two intervention paradigms: The static prompt intervention paradigm can be conceptualized as an interface for non-AI experts (e.g., domain specialists or cognitive science researchers) to inject their domain expertise and cognitive theories into model reasoning through rule-based interventions. This paradigm specifically enables specialists across domains to create targeted PI designs adapted to their unique scenarios. The dynamic prompt intervention paradigm empowers LRMs to evolve into superior meta-thinkers capable of adapting their reasoning strategies across diverse scenarios during test-time. To further advance this direction, we encourage deeper investigation into LRMs' reasoning behaviors, which could refine any module (when-how-which) of the proposed framework. Such developments would enable more effective reasoning interventions, ultimately unlocking the full potential of test-time computation. 

In addition to optimizing the reasoning behaviors during testing, our framework can be seamlessly adapted to the trajectory sampling phase during the training of reinforcement learning. Through prompt interventions, it enables the collection of finer-grained and more diverse trajectories. This allows models to directly internalize carefully designed test-time prompt intervention patterns, a promising direction we reserve for future work.

\bibliography{aaai2026}

\definecolor{awesome}{rgb}{1.0, 0.13, 0.32}
\definecolor{azure(colorwheel)}{rgb}{0.0, 0.5, 1.0}
\definecolor{aureolin}{rgb}{0.99, 0.93, 0.0}
\definecolor{amber}{rgb}{0.99, 0.93, 0.0}
\definecolor{frenchrose}{rgb}{0.96, 0.29, 0.54}
\definecolor{coquelicot}{rgb}{1.0, 0.22, 0.0}
\definecolor{aliceblue}{rgb}{0.9, 0.9, 0.9}

\newpage

\appendix

\section{Preliminary Experimental Setups.}
For experiment in Figure 2(a), step-level attention scores are computed by averaging attention values across tokens in the current step with respect to tokens in each preceding step, with the resulting attention maps presented in Figure 2(a). To mitigate the influence of attention sinks, we mask out attention contributions from the first and last three token positions in the sequence, and then normalize the remaining attention values. For the 36-layer Qwen3-8B model, token-level attention is averaged across higher layers (31 to 35) and all attention heads.


%

For experiment in Figure 3(a), we gathered the raw responses generated by Qwen3-8B and DeepSeek-R1-Distill-Qwen-14B on the MATH-500 and GPQA datasets, employing word cloud visualization to represent word frequency distributions, with word size corresponding to occurrence frequency. Common function words were filtered out, including articles ("the"), prepositions ("on", "for"), conjunctions ("and"), pronouns ("it"), and other similar terms.

For experiment in Figure 3(b) and 3(c), we segmented the reasoning trajectories using '\textit{\textbackslash n\textbackslash n}' as delimiters and classified each step based on whether it contained verification-related phrases, such as "wait", "let me verify", "let me check", "checking", "verifying", and "double-check".

For experiment in Figure 3(d), We carried out this investigation using Qwen3-8B, with the model’s native reasoning outputs serving as the baseline. In the \texttt{NoVerify} setting, we extracted the thinking content from the baseline, masked all verification steps as well as steps containing answers, and then prompted the model to summarize and generate a conclusion by concatenating with end-of-thinking delimiter $\langle\textit{/think}\rangle$.
In the \texttt{NoWait} setting, we replaced the trigger word "\textit{Wait}" with "\textit{So}" during generation to discourage unnecessary verification steps.

\section{More analyses for Word Clouds.}

Figure 3(a) in the main text presents a word cloud visualization of the original thoughts from Qwen3-32B on OlympiadBench. In addition, we provide word clouds for additional models across multiple benchmarks, as illustrated in Figures 8, 9, 10, and 11. Through examination of these visualizations, we observe that current reasoning models exhibit similar patterns of excessive self-verification. While medium-probability tokens correlate with dataset-specific characteristics, the most frequently occurring high-probability tokens consistently include terms such as "check," "wait," and similar expressions. This finding further corroborates our hypothesis presented in the main text.

\begin{figure}[!t]
  \centerline{\includegraphics[scale=0.26]{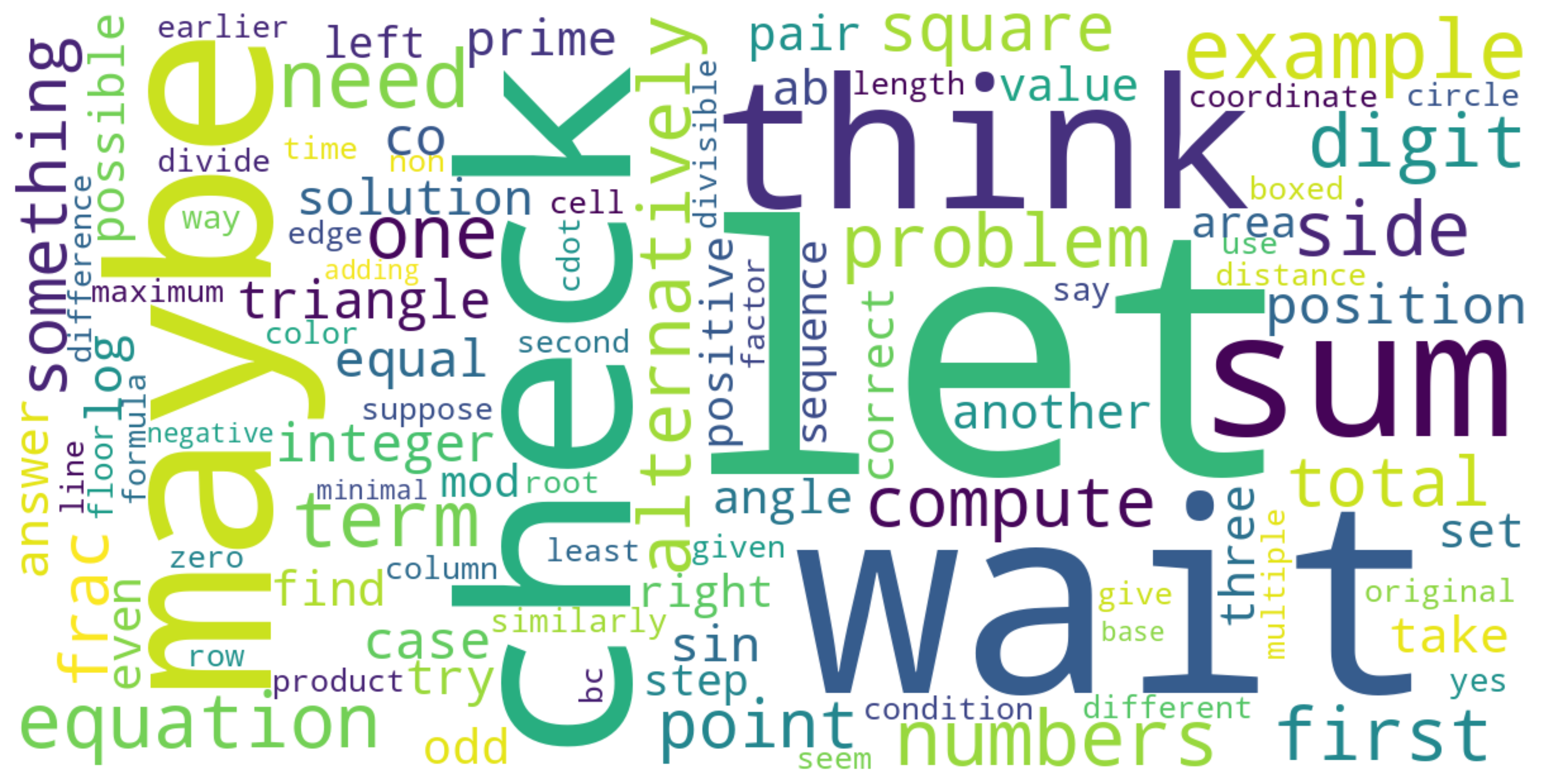}}
  \caption{The word cloud visualization of the Qwen3-32B’ CoTs on the MATH-500.}
  \label{figure5}
\end{figure} 

\begin{figure}[!t]
  \centerline{\includegraphics[scale=0.26]{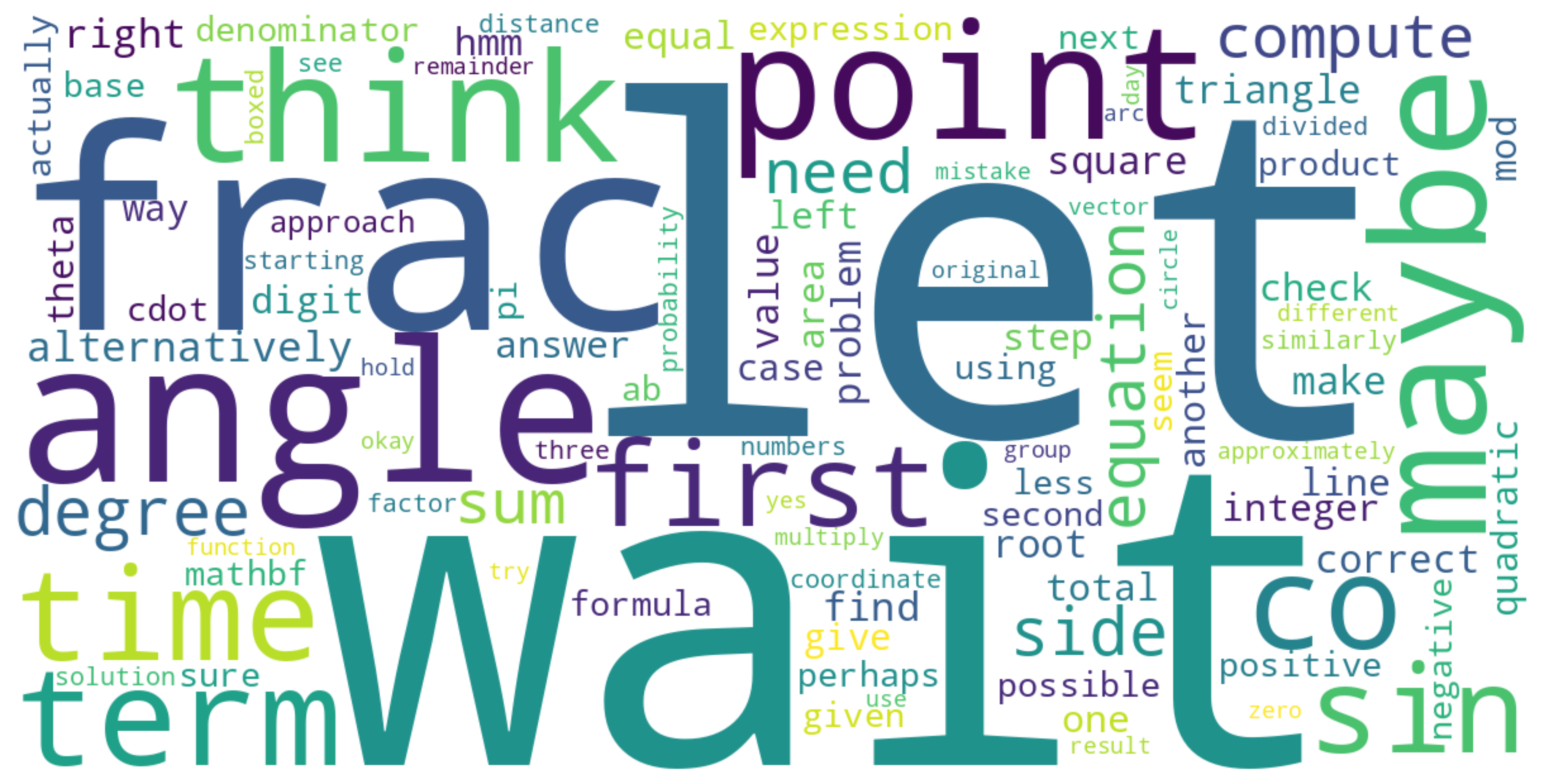}}
  \caption{The word cloud visualization of the DeepSeek-R1-Distill-Qwen-32B’ CoTs on the MATH-500.}
  \label{figure5}
\end{figure} 

\begin{figure}[!t]
  \centerline{\includegraphics[scale=0.26]{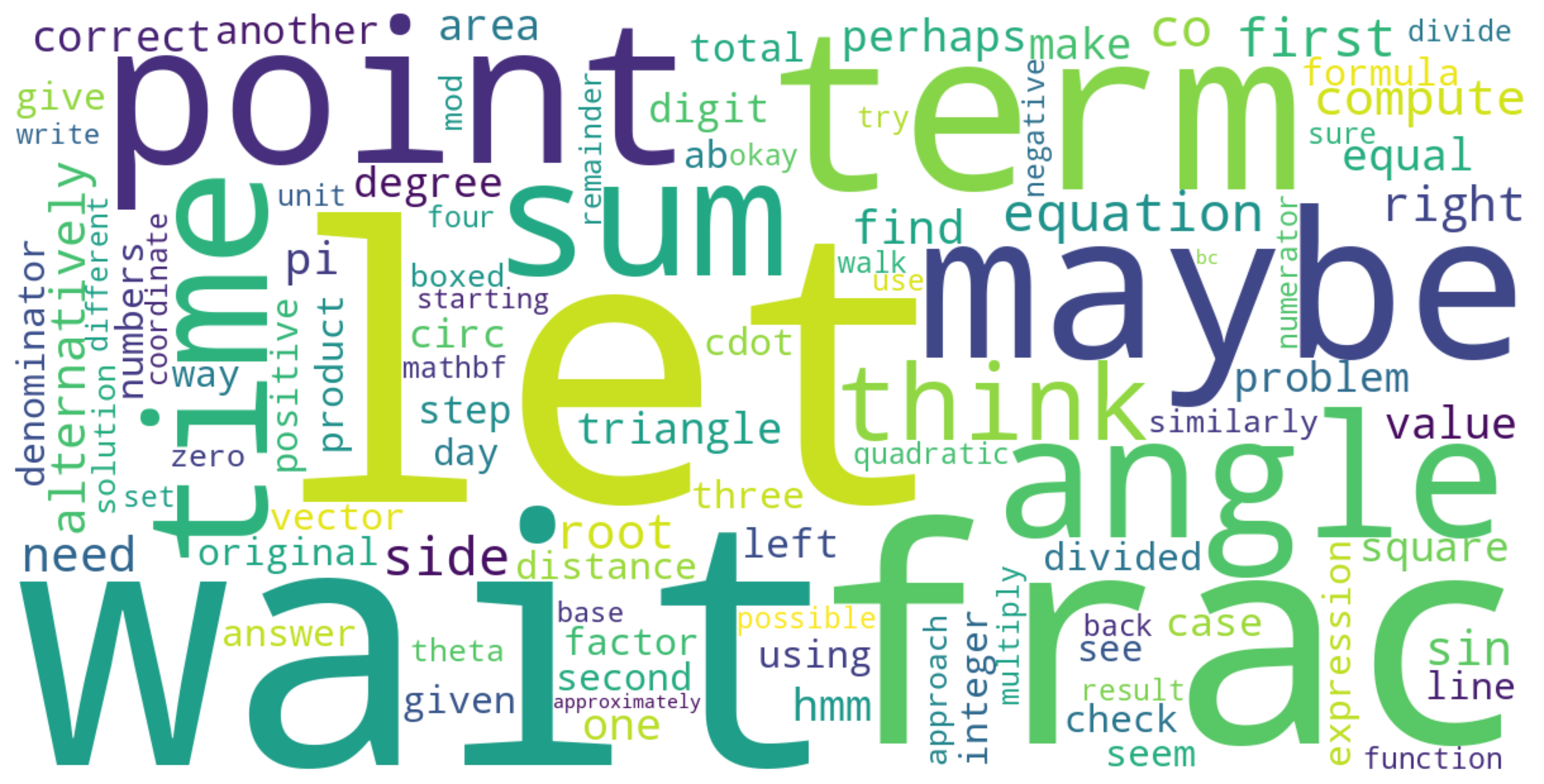}}
  \caption{The word cloud visualization of the DeepSeek-R1-Distill-Qwen-7B’ CoTs on the MATH-500.}
  \label{figure5}
\end{figure} 

\begin{figure}[!t]
  \centerline{\includegraphics[scale=0.26]{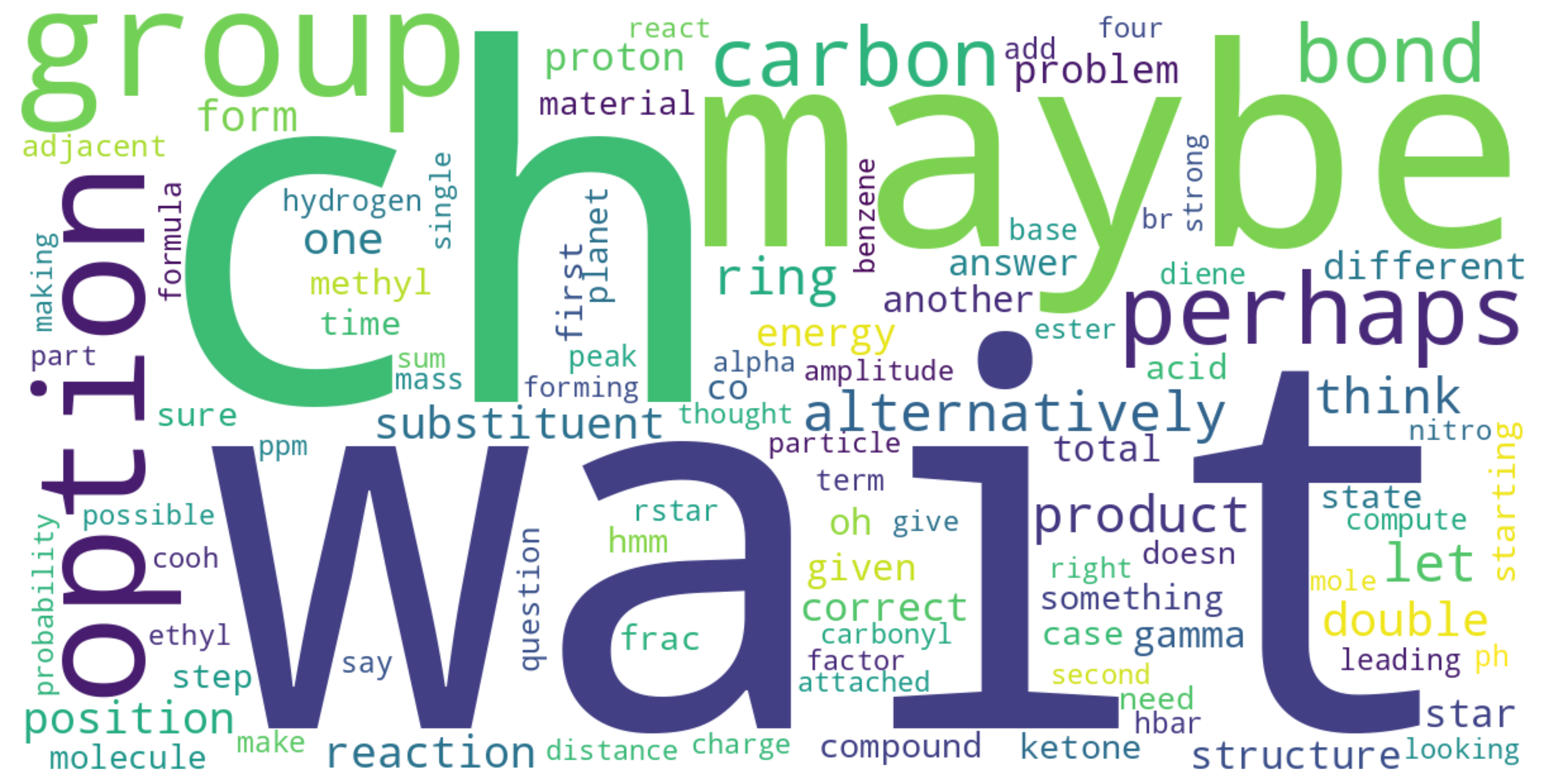}}
  \caption{The word cloud visualization of the DeepSeek-R1-Distill-Qwen-7B’ CoTs on the GPQA.}
  \label{figure5}
\end{figure} 

\section{Detailed analysis for Methodology.}

\subsection{Indicator of Reasoning Depth}
\paragraph{Illustrative Example.} In Figure 8, we illustrate through a case study why the JSD difference between early and final layer probability distributions can represent reasoning depth. The figure shows heatmaps of JSD values across different layers for two text segments. The upper plot demonstrates that when the text content involves shallow information processing such as self-affirmation reflection or repetition of previously generated content, the distributional differences between high and low layers are minimal. Conversely, the lower plot shows that when the text content involves deep thinking such as logic-based mathematical reasoning, the distributional differences between high and low layers are substantial. This intuitively explains how RDS facilitates efficient reasoning.

\paragraph{Theoretical Justification.} Intuitively, the examples suggest a correlation between reasoning processes and their corresponding depth scores. Here, we provide a rigorous theoretical analysis by mapping the abstract concept of reasoning depth to a measurable geometric quantity.

We conceptualize the hidden state space of an $L$-layer Transformer as a high-dimensional \textbf{semantic manifold} $\mathcal{M}$. For a given reasoning task, as the model processes information layer by layer, its sequence of hidden states $\{h^{(1)}, h^{(2)}, \dots, h^{(L)}\}$ traces a representational trajectory $\mathcal{T}$ on this manifold. Therefore, the layer-wise computation process of Transformers can be viewed as a dynamic geometric trajectory that performs non-trivial geometric transformations on representations. Within this framework, information processing can be categorized into deep information transformation and shallow information propagation. The former corresponds to long, highly nonlinear segments of trajectory $\mathcal{T}$, characterized by large representational displacements and high curvature, indicating that information undergoes complex twisting and folding on the manifold. In contrast, the latter corresponds to short, approximately linear segments of trajectory $\mathcal{T}$, with small representational displacements and gradual directional changes, suggesting that information is merely translated on the manifold without fundamental semantic transitions.

Let $\Delta h^{(l)} = h^{(l+1)} - h^{(l)}$ denote the representational displacement vector at layer $l$. We can characterize the geometric transformation of representations by computing the arc length of this trajectory in the hidden space, thereby quantifying the abstract concept of reasoning depth:
$$\text{ArcLength}(\mathcal{T}) = \sum_{l=1}^{L-1} ||\Delta h^{(l)}||_2,$$
where $L$ denotes model layers, and $h$ denotes hidden states.

However, computing complete trajectory geometry for each token directly in high-dimensional hidden spaces is computationally prohibitive. Moreover, due to representation space anisotropy, Euclidean distances in hidden spaces fail to consistently correspond to semantic distances. To overcome these challenges, rather than measuring directly in the complex hidden space, we project the trajectory onto a more structured space.

Specifically, we define the language model head and softmax function together as a probability projection operator $\Phi: \mathcal{M} \to \Delta^{|V|-1}$. This operator maps internal states $h^{(l)}$ in the thinking space $\mathcal{M}$ to external preferences $p^{(l)}$ in the decision space (the probability simplex $\Delta^{|V|-1}$ over the vocabulary).
$$
p^{(l)} = \Phi\left(h^{(l)}\right) = \operatorname{softmax}\left(W_{LM} \cdot h^{(l)}\right)
$$

Through this projection, the complex high-dimensional representation trajectory $\mathcal{T}$ is transformed into a lower-dimensional probability trajectory $\mathcal{T}_p$.

Thus, we can compute chord lengths along the projected trajectory using JSD, serving as a proxy for reasoning depth that effectively quantifies the intensity of the unobservable thinking process in hidden space:
$$\text{JSD}(p^{(l)} || p^{(L)}) \approx c \cdot ||p^{(l)} - p^{(L)}||_2^2$$

%



\subsection{High-entropy Intervention}
\paragraph{Illustrative Example.}
Figure 11 demonstrates why we design interventions at high-entropy points. When the model generates the first token of a step (the action trigger token) with very low entropy, such as 0 as shown in the figure, it has already implicitly determined the subsequent reasoning path. Forced intervention at this point often results in low-quality content that may not follow the intervention prompt. In the figure, the content generated after the summary intervention trigger is not actually summarization content, and the text contains multiple instances of self-checking with verbose output. Therefore, intervening only at high-entropy points prevents resource waste.

\paragraph{Theoretical Justification.}
We model the reasoning generation process as a Partially Observable Markov Decision Process (POMDP), where:
The state space $\mathcal{S}$ represents a collection of world states not directly observable by the model.
The action space $\mathcal{A}$ denotes the set of actions available to the reasoning model.
The reward function $R(s, a)$ specifies the immediate reward obtained after executing action $a$ in state $s$.
The discount factor $\gamma \in [0,1]$ balances immediate and future rewards.

We define the action-value function (Q-function) $Q^\pi(b, a)$ as the expected cumulative reward for executing action $a$ under belief $b$ and then following policy $\pi$:
$$Q^\pi(b, a) = \mathbb{E}_{\pi} \left[ \sum_{k=0}^{\infty} \gamma^k R_{t+k} \Big| b_t=b, a_t=a \right]$$
The state-value function (V-function) $V^\pi(b)$ represents the expected Q-value under belief $b$ when following policy $\pi$:
$$V^\pi(b) = \mathbb{E}_{a \sim \pi(\cdot|b)} [Q^\pi(b, a)] = \sum_{a \in \mathcal{A}} \pi(a|b)Q^\pi(b, a)$$

Within this framework, we define an intervention $I$ as an external operation that replaces the model's original stochastic policy $\pi$ with a specified, typically deterministic new policy $\pi_I$. For example, an intervention forcing "\textbf{Progression}" would have policy $\pi_I(a_p)=1$, where $a_p$ represents the "\textbf{Progression}" tokens.

We define the Value of Intervention (VoI) as the difference in value functions between post-intervention and pre-intervention states under the current belief state.
For an intervention $I$ with policy $\pi_I$ at belief state $b$, the value gain is defined as:
$$\text{VoI}(b, I) = V^{\pi_I}(b) - V^\pi(b)$$

Since our interventions typically select a deterministic action $a_I$, we have $V^{\pi_I}(b) = Q^\pi(b, a_I)$. Substituting this expression, we obtain the VoI formula:
$$\text{VoI}(b, I) = Q^\pi(b, a_I) - \sum_{a \in \mathcal{A}} \pi(a|b)Q^\pi(b, a)$$

Intuitively, this formula states that intervention value equals the value gained from forcing a good action minus the average value from the model's uncertain exploration. Next, we analyze how VoI varies with policy entropy $H(\pi)$.

\textbf{Case 1: Low entropy state ($H(\pi) \to 0$).} When entropy is low, the model's policy distribution $\pi(\cdot|b)$ becomes highly peaked. This implies the existence of an optimal action $a^*$ such that $\pi(a^*|b) \approx 1$, while for all other actions $a \neq a^*$, we have $\pi(a|b) \approx 0$.

In this case, the model's original value function becomes:
\begin{equation*}
\begin{split}
V^\pi(b) &= \sum_{a \in \mathcal{A}} \pi(a|b)Q^\pi(b, a) \\
         &\approx 1 \cdot Q^\pi(b, a^*) + \sum_{a \neq a^*} 0 \cdot Q^\pi(b, a) \\
         &= Q^\pi(b, a^*)
\end{split}
\end{equation*}

Substituting into the VoI expression, we obtain:
$$\text{VoI}(b, I) \approx Q^\pi(b, a_I) - Q^\pi(b, a^*)$$

Since $a^*$ is the action chosen by the model with extremely high confidence, it is likely the truly optimal action under the current belief, meaning $Q^\pi(b, a^*) \geq Q^\pi(b, a_I)$ for any intervention action $a_I$ we specify. Therefore:
$$\text{VoI}(b, I) \leq 0$$

In low entropy states, intervention yields negative or zero expected value gain. Intervention at this point is not only redundant but may disrupt the model's optimal decision path, causing performance degradation.

\textbf{Case 2: High entropy state ($H(\pi) \to \log_2 k$).}
When entropy is high, the model's policy distribution $\pi(\cdot|b)$ approximates a uniform distribution over a subset of $k$ actions $\mathcal{A}' = \{a_1, \ldots, a_k\}$, where $\pi(a_i|b) \approx 1/k$ for all $a_i \in \mathcal{A}'$.

This represents a decision crossroads where the model perceives these $k$ actions as having comparable plausibility or value, unable to make a clear choice. The model's original value function becomes:
$$V^\pi(b) \approx \frac{1}{k} \sum_{i=1}^{k} Q^\pi(b, a_i)$$

Our intervention strategy $I$ selects the action we deem highest-value, such as $a_p$ (Progression), so $a_I = a_p$. Substituting into the VoI expression:
\begin{equation*}
\begin{split}
\text{VoI}(b, I) &\approx Q^\pi(b, a_p) - \frac{1}{k} \sum_{i=1}^{k} Q^\pi(b, a_i) \\
                &\approx \frac{k-1}{k} \left( Q^\pi(b, a_p) - \frac{1}{k-1} \sum_{\substack{a_i \in \mathcal{A}' \\ a_i \neq a_p}} Q^\pi(b, a_i) \right)
\end{split}
\end{equation*}

This formula shows that intervention value (VoI) is proportional to the difference between the chosen optimal action's value and the average value of all other candidate actions.

At high-entropy crossroads, the candidate action set $\mathcal{A}'$ likely contains both high-value correct paths (like $a_p$) and numerous low-value redundant, exploratory, or erroneous paths (like $a_v$: verification, $a_e$: exploration). This means $Q^\pi(b, a_p)$ significantly exceeds the average Q-value of other actions. Therefore:
$$\text{VoI}(b, I) > 0$$

In high entropy states, intervention yields significantly positive expected value gain. Higher entropy means the average value of candidate actions is increasingly diluted by low-value actions. Consequently, forcing high-value actions produces a stronger purification effect, resulting in greater value gain.

In summary, intervention during low entropy states is high-risk with low returns. However, at high-entropy decision crossroads, external intervention provides maximum value gain, helping the model prune suboptimal exploration branches and achieve more efficient and reliable reasoning.

\subsection{Detailed Illustration of PI}
In Figure 10, we demonstrate the specific process of one Prompt Intervention. At step 15, PI first computes the entropy of the first token generated in the step. Since the entropy exceeds the threshold of 0.3, the How module intervenes to generate multiple branches. Subsequently, after obtaining sequence probabilities and RDS, the Which module compares the combined scores. As shown in the figure, the progressive branch achieves the highest score and is therefore selected as the current reasoning step. Both human intervention branches provide correct answers and generate thinking termination delimiter $\langle\textit{/think}\rangle$, while the model's spontaneous branch continues attempting additional methods to solve an already correctly answered problem, exhibiting overthinking issues. This example demonstrates that PI effectively alleviates such problems.




\section{More Experimental Setups.}
\subsection{More Benchmark Details.}

\paragraph{\textsc{Mathematical Evaluation Benchmarks}}.
To comprehensively assess the performance of models on mathematical reasoning tasks, we employ a diverse set of benchmark datasets that span various levels of difficulty, from elementary arithmetic to advanced competition-level mathematics.

\begin{itemize}
\item \textbf{GSM8K} is a carefully curated dataset consisting of 1,319 grade-school math problems. It is specifically designed to evaluate the ability of models to perform multi-step reasoning in foundational mathematical tasks. Each problem typically requires between two and eight sequential operations, relying primarily on basic arithmetic and demanding accurate handling of intermediate results.

\item \textbf{MATH-500} is a challenging collection of high-school-level problems drawn from multiple domains, including Prealgebra, Algebra, and Number Theory. These problems are typically sourced from competitive mathematical contests and require abstract thinking and complex logical deduction. To ensure comparability with prior work, we adopt the standard 500-problem subset originally curated by OpenAI for evaluation.

\item \textbf{AMC 2023} is composed of 40 problems from the 2023 edition of the American Mathematics Competitions (AMC), organized annually by the Mathematical Association of America (MAA). The competition aims to foster problem-solving skills and identify mathematical talent among students. The selected problems cover key areas such as algebra, geometry, number theory, and combinatorics, serving as a robust testbed for evaluating advanced mathematical reasoning.

\item \textbf{OlympiadBench} is a bilingual, multimodal benchmark dataset at the Olympiad level, designed to challenge and evaluate the advanced reasoning capabilities of large language models and multimodal systems. It includes a total of 8,476 problems from mathematics and physics competitions, including those from the Chinese Gaokao. In our experiments, we use the same subset of 675 questions employed in LIMO, enabling rule-based evaluation of generated responses.

\end{itemize}

\paragraph{\textsc{STEM Evaluation Benchmarks.}}
To evaluate the models' performance on advanced scientific reasoning tasks, we utilize two specialized benchmark datasets that cover both undergraduate-level and expert-level science problems.

\begin{itemize}
 \item \textbf{GPQA} is a rigorously constructed benchmark featuring high-quality questions across disciplines such as physics, chemistry, and biology. A notable characteristic of this dataset is its difficulty: even domain experts with PhDs achieved only 69.7\% accuracy during initial evaluations. For our experiments, we use the most refined and reliable subset of the dataset, known as GPQA Diamond, which consists of 198 carefully vetted questions.

\item \textbf{Minerva} is a collection of undergraduate-level STEM problems requiring multi-step reasoning, primarily drawn from university courses such as “solid-state chemistry,” “information and entropy,” “differential equations,” and “special relativity.” Each problem is designed to yield an automatically verifiable solution—either numerical or symbolic, often checked using SymPy. These questions have been reformulated by human annotators to be self-contained and to feature a clearly identifiable final answer. Problems requiring proofs or open-ended responses were excluded. In total, we selected 272 problems for evaluation, of which 191 have numeric solutions and 81 admit symbolic verification.
\end{itemize}

\paragraph{\textsc{Hallucination Benchmarks.}} To evaluate model susceptibility to reasoning errors that produce factually or logically incorrect outputs despite seemingly coherent internal reasoning, we employ two targeted benchmarks designed to assess truthfulness and robustness in question answering.
\begin{itemize}
\item \textbf{TruthfulQA} is a benchmark designed to evaluate the factual accuracy of language models when answering complex, real-world questions. It comprises 817 questions across 38 categories, including health, law, finance, and politics. Many questions are crafted to expose common misconceptions or widely held false beliefs, making the benchmark particularly effective at identifying whether a model favors popular but inaccurate responses over truthful ones. For this benchmark, we retain the two evaluation methods from the original work: Multiple Choice 1 (MC1) and Multiple Choice 2 (MC2), where MC1 has only one correct answer, while MC2 has multiple correct answers and answering any one correctly is sufficient.
\item \textbf{GSM-NoOp} is a modified version of the GSM8K dataset constructed to test model resilience to reasoning hallucinations. In this benchmark, semantically irrelevant yet contextually plausible "no-op" phrases are inserted into mathematical word problems. While these additions do not affect the correct reasoning path, they can mislead models that rely on superficial pattern recognition rather than deep logical understanding. This design allows GSM-NoOp to specifically probe for reasoning inconsistencies—instances where the internal thought process appears valid but leads to an incorrect final answer. Following the methodology outlined in \cite{GSMnoopmirzadeh2024gsmsymbolicunderstandinglimitationsmathematical,sun2025detectionmitigationhallucinationlarge}, we sample 1,000 examples from the GSM8K dataset. For each question, we use GPT-4.1 \cite{openai2024gpt4technicalreport} to generate a no-op phrase using the prompt detailed in Figure 12, then integrate the generated phrase into the original question using another GPT-4.1 call following the template in Figure 13.
\end{itemize}


\subsection{More Metric Details.}
For evaluation, we employ three metrics: \textit{Accuracy} (\textbf{Acc}), \textit{Token Number} (\textbf{Tok}), and \textit{Compression Rate} (\textbf{CR}). 
\textbf{Acc} measures the final answer accuracy. \textbf{Tok} represents the average generation length per sample, serving as a proxy for computational cost. \textbf{CR} is defined as the ratio of the average response length to that of the original model, where lower values indicate higher compression efficiency. We conduct 4 sampling rounds per instance and average the results across all metrics to ensure stability and reliability.

\subsection{More Baseline Details.}

\textit{Vanilla} performs direct evaluation of the LRLM without any intervention. \textit{NoThinking} prompts the model to skip the reasoning phase and generate the final answer directly. \textit{NOWAIT} disables explicit self-reflection by suppressing tokens such as 'Wait' and 'Hmm' during inference. \textit{DEER} employs self-truncation of CoT sequences through early exit during generation when the model demonstrates high confidence in a trial answer.

\subsection{More Implementation Details.}
We implemented PI using HuggingFace Transformers. All evaluations are conducted in a zero-shot Chain-of-Thought (CoT) setting with the following prompt: \textit{"Please reason step by step, separate logical reasoning steps with two newline characters (\textbackslash n\textbackslash n), keep each reasoning step within approximately 100 tokens, and put your final answer within \textbackslash boxed\{\}."} For the decoding strategy, we employ top-p sampling with the officially recommended parameters of $temperature=0.6$
 and $p=0.95$. Since the ground-truth answers to all evaluation problems in our experiments are well-structured numerical values or options, we apply rule-based evaluation directly to verify equivalence. We set the maximum generation length to 16,384, and set the hyperparameter $\alpha$ to $0.6$. To mitigate the effects of randomness, the results presented in the table represent the average of eight experimental runs. For the early layers used to compute JSD, we selected three layers to compare against the final layer: the third-to-last, seventh-to-last, and eleventh-to-last layers, with a 4-layer interval between each selection. Specifically, for DeepSeek-R1-Distill-Qwen-14B (48 layers total), we used layers 46, 42, and 38; for Qwen3-4B (36 layers total), layers 34, 30, and 26; for Qwen3-8B (36 layers total), layers 34, 30, and 26; and for Qwen3-14B (40 layers total), layers 38, 34, and 30. We conducted experimental investigation on this hyperparameter selection, and the results indicate that layer choice has minimal impact on performance. For the high-entropy intervention design in the When module, we compute the initial token's entropy using the first four tokens. The entropy threshold for intervention timing is set to 0.3, where intervention occurs when entropy exceeds this threshold. All experiments were run on the NVIDIA H20 96G. For the intervention operations, we employed the following prompts in our experiments to implement the respective reasoning phases:
 \begin{itemize}
    \item \textbf{Progression}: \textit{"Okay, moving on."}
    \item \textbf{Summary}: \textit{"So, putting it all together"}
    \item \textbf{Verification}: \textit{"Wait, let me verify."}
    \item \textbf{Conclusion}: \textit{"**Final Answer**\textbackslash n\textbackslash boxed"}
 \end{itemize}




\section{More Experimental Results.}
\subsection{Main Results and Analysis.}
\paragraph{Complete Results.}
As shown in Table 1, we report experimental results across different model sizes, where PI consistently exceeds baseline performance, attaining the highest accuracy with minimal generation overhead. We provide a more detailed analysis of the experimental results in comparison with each baseline method.

\paragraph{Comparison with Baselines.} We conduct a detailed comparison of the improvements of PI over each baseline and attempt to analyze the underlying reasons. Specifically, while \textit{NoThinking} trades off substantial accuracy for CoT length reduction, PI simultaneously reduces reasoning length and enhances accuracy. Relative to \textit{NOWAIT}, PI achieves greater reductions in reasoning length due to fundamental differences in approach. \textit{NOWAIT} employs a passive blacklist mechanism that suppresses tokens potentially associated with overthinking, but this strategy suffers from limited coverage—models can readily circumvent such restrictions through alternative phrasings. In contrast, PI addresses overthinking at its source by intervening in reasoning patterns and teaching the model more effective thinking strategies. Compared to \textit{DEER}, PI exhibits more consistent CoT compression performance across benchmarks of varying complexity and maintains effectiveness even on challenging problems. Additionally, PI's reasoning paths are more complete, avoiding the readability degradation issues caused by early exit (the thought process could be truncated).

\paragraph{Performance on More LRMs.} In addition to the four models mentioned in the main text, we additionally supplemented experiments on DeepSeek-R1-Distill-Llama-8B and DeepSeek-R1-Distill-Qwen-7B. Among these, the experiment on DeepSeek-R1-Distill-Llama-8B was conducted to demonstrate that PI remains effective under different model architectures and pretraining data. We visually present the performance improvements of PI compared to Vanilla CoT across 6 models in the form of a bar chart in Figure 9. The experimental results demonstrate that the effectiveness of PI is universal and robust, achieving higher accuracy with shorter reasoning lengths across all 6 models.

\subsection{More Ablation Results.}
We present all ablation experimental results in Table 3, including OlympiadBench and Minerva. Furthermore, we provide detailed calculations of compression rates and average results across six datasets. The experimental findings demonstrate that the conclusions analyzed in the main text remain valid across these additional datasets.
Additionally, we conducted ablation experiments on trigger words, with results shown in Table 5. The results demonstrate that PI is robust to trigger words, and we encourage boldly setting different predefined trigger words for different intervention actions across various task scenarios.


\section{More Cases.}

Figure 14 illustrates another example of efficient reasoning. The original generation attempts to verify conclusions by calculating specific numerical values during problem-solving, leading to lengthy and ineffective thinking, while PI achieves the correct conclusion through more concise reasoning. Figure 15 presents another example of hallucination mitigation. Through additional verification and reflection, PI identifies more pathways through which Visionaries in California gain inspiration and ultimately recognizes that the sources of inspiration for Visionaries in California are diverse, leading to the correct option. Conversely, the baseline model rushes to conclusions after recalling only a single inspirational pathway for Visionaries in California, leading to an incorrect response.

\begin{table}[ht]
\begin{tabular}{|ccccc|}
\hline
\multicolumn{5}{|c|}{\textbf{GPQA-Diamond}} \\ \hline
\multicolumn{1}{|c|}{} & \multicolumn{1}{c|}{Res-Tok} & \multicolumn{1}{c|}{Gen-Tok} & \multicolumn{1}{c|}{Lat.} & Freq. \\ \hline
\multicolumn{1}{|c|}{Vanilla} 
& \multicolumn{1}{c|}{9105} & \multicolumn{1}{c|}{9105} 
& \multicolumn{1}{c|}{715s} & -- \\ \hline
\multicolumn{1}{|c|}{PI} 
& \multicolumn{1}{c|}{\textbf{3986 (44\%)}} & \multicolumn{1}{c|}{\textbf{7045 (77\%)}} 
& \multicolumn{1}{c|}{\textbf{339s}} & 52.0\% \\ \hline
\multicolumn{5}{|c|}{\textbf{MATH-500}} \\ \hline
\multicolumn{1}{|c|}{} & \multicolumn{1}{c|}{Res-Tok} 
& \multicolumn{1}{c|}{Gen-Tok} 
& \multicolumn{1}{c|}{Lat.} & Freq. \\ \hline
\multicolumn{1}{|c|}{Vanilla} 
& \multicolumn{1}{c|}{5224} & \multicolumn{1}{c|}{5224} 
& \multicolumn{1}{c|}{278s} & -- \\ \hline
\multicolumn{1}{|c|}{PI} 
& \multicolumn{1}{c|}{\textbf{3013 (58\%)}} & \multicolumn{1}{c|}{\textbf{4710 (90\%)}} 
& \multicolumn{1}{c|}{\textbf{209s}} & 44.3\% \\ \hline
\end{tabular}
\caption{Computational cost comparison of Vanilla CoT and PI on two benchmarks. Experiments are conducted on Qwen3-8B.}
\label{cost}
\end{table}

\section{Computational Cost Analysis.}
\subsection{Experimental Analysis}
In this section, we conduct a comprehensive analysis of the computational cost savings achieved by PI. We conduct a comprehensive comparison from multiple perspectives including response token cost (Res-Tok), total generated token cost (Gen-Tok), memory usage, and generation latency (Lat.), with experimental results shown in Table 4.
The table shows performance comparisons on a simpler benchmark (MATH-500) and a more challenging one (GPQA). On the more difficult GPQA, PI's average generation latency per sample is 47\% of the baseline method, demonstrating significant latency savings. While PI's simultaneous multi-branch generation slightly slows single token generation and the scoring computation in the Which module introduces additional latency, these overheads are negligible compared to the latency reduction from shortened sequences, since attention computation scales quadratically with sequence length.
We also measured PI's total token generation cost by including tokens from unselected branches, which amounts to 77\% of the baseline. Thanks to the high-entropy intervention strategy in the How module, PI intervenes only 55\% of the time, making the token waste acceptable. We believe that future effective static intervention strategies could eliminate token waste entirely, further reducing inference time costs.
For the simpler MATH-500 benchmark, since baseline generation lengths are already short, PI's latency reduction is modest. However, PI still effectively improves accuracy while reducing response tokens and enhancing model interpretability. 
Regarding {memory usage}, the additional memory introduced by PI is negligible due to reusable key-value cache from generated steps, while its sequence length reduction effectively saves key-value cache memory overhead. Our peak memory measurements show the baseline consuming 46,834MB compared to PI's 32,454MB.
Based on this analysis, we believe PI truly achieves efficient reasoning.

\subsection{Theoretical Analysis}
In addition to empirical overhead measurements, we provide a theoretical analysis to demonstrate that PI effectively reduces computational costs. Let $L$ denote the total length generated by the original CoT method, and $\alpha$ represent PI's compression ratio relative to $L$, such that PI generates a sequence of length $\alpha L$. Let $s$ be the total number of steps in the original CoT reasoning, and $\beta$ be the proportion of steps where PI identifies high entropy and performs branching, yielding $\alpha\beta s$ as the number of branching steps in PI. During transformer inference, the primary computational overhead stems from attention calculations, which constitutes our main focus. Assuming the generation process employs key-value caching technology, each new token only needs to compute attention with the cached key-value pairs.

For the original CoT method, the computational cost is:
\begin{equation}
T=O(1)+O(2)+\dots + O(L)=O(L^2)
\end{equation}

For our PI, considering the case of three branches, the computational cost comprises two components: the main path length (the length of the actually displayed response) and the additional overhead from discarded branch steps (two additional branches). We calculate each component separately.

First, we calculate the cost of the main path:
\begin{equation}
T_{\text{main}} = \sum_{t=1}^{\alpha L} t = \frac{\alpha L (\alpha L + 1)}{2} = O\left(\alpha^{2} L^{2}\right) 
\end{equation}

Then, we calculate the additional overhead from discarded branch steps:

For the i-th branching step, we generate 3 branches, each of length $\frac{L}{s}$, where 1 branch is part of the main path. Therefore, we need to compute 2 additional branches.

Computation for a single additional branch:
Let the prefix length be $p_i = i \times \frac{L}{s}$, and the branch length be $l = \frac{L}{s}$.
The computational cost for generating this branch is:
\begin{equation}
 C_i =\sum_{j=1}^{l}\left(p_{i}+j\right) = l \cdot p_{i} + \sum_{j=1}^{l} j = l \cdot p_{i} + \frac{l(l+1)}{2} 
\end{equation}

Based on this, we calculate the total overhead for all additional branches. Since there are $\alpha\beta s$ branching steps in total, each producing 2 additional branches, the overhead for additional branches is:
\begin{equation}
\begin{aligned}
T_{\text{add}} &= \sum_{i=1}^{\beta\alpha s} 2 \times \left[ \frac{L}{s} \times i \times \frac{L}{s} + \frac{L/s \ (L/s + 1)}{2} \right] \\
&= \sum_{i=1}^{\beta\alpha s} 2 \times \left[ \frac{i L^{2}}{s^{2}} + \frac{L^{2}/s^{2} + L/s}{2} \right]
\end{aligned}
\end{equation}

Simplifying the first term in Equation 7:
\begin{equation}
\begin{aligned}
\sum_{i=1}^{\beta\alpha s} \frac{2i L^{2}}{s^{2}} 
&= \frac{2 L^{2}}{s^{2}} \times \frac{\beta\alpha s (\beta\alpha s + 1)}{2} \\
&\approx \frac{(\beta\alpha)^{2} L^{2} s^{2}}{s^{2}} 
= (\beta\alpha)^{2} L^{2}
\end{aligned}
\end{equation}

Simplifying the second term in Equation 7:
\begin{equation}
\sum_{i=1}^{\beta\alpha s} \frac{L^{2}/s^{2} + L/s}{1} = \beta\alpha s \times \frac{L^{2}/s^{2} + L/s}{1} \approx \frac{\beta\alpha L^{2}}{s}
\end{equation}

When $s$ is large, the second term becomes negligible, therefore:
\begin{equation}
T_{\text{add}} = O\left((\beta\alpha)^{2}L^{2}\right)
\end{equation}

The total cost of PI is:
\begin{equation}
\begin{aligned}
T_{\mathrm{PI}} &= T_{\text{main}} + T_{\text{add}} \\ 
&=
O\left(\alpha^{2}L^{2}\right) + O\left((\beta\alpha)^{2}L^{2}\right) \\
&= O\left(\alpha^{2}L^{2}\left(1+\beta^{2}\right)\right)
\end{aligned}
\end{equation}

We can calculate the computational cost savings ratio of PI compared to the original CoT as:
\begin{equation}
Saving Ratio= 1 - \frac{T_{PI}}{L^{2}} = 1 - \alpha^{2}\left(1 + \beta^{2}\right)
\end{equation}

Based on the experimental results, we can conclude that both the compression ratio and the proportion of high-entropy steps are approximately $0.5$. Substituting $\alpha = 0.5$ and $\beta = 0.5$, we obtain:
\begin{equation}
Saving Ratio = 1 - 0.5^{2}\left(1 + 0.5^{2}\right)=68.75\%
\end{equation}

The above represents the upper bound of computational cost savings we have calculated, as it assumes branches occur as early as possible in the reasoning process. Below, we compute a more complex but realistic overhead estimate:

We assume that branching positions are uniformly distributed throughout the reasoning path. That is, PI comprises a total of $\alpha s$ steps, among which $\alpha\beta s$ steps involve branching. These branching steps are uniformly distributed across the entire path, with branching positions occurring at:
\begin{equation}
\left\{
\frac{k \cdot \alpha s}{\beta \alpha s} = \frac{k}{\beta} 
\, : \, 
k = 1, 2, \ldots, \beta\alpha s
\right\}
\end{equation}

In this scenario, the computational cost of the main path remains unchanged at $O(\alpha^2 L^2)$. For the computational cost of a single additional branch, since the k-th branch occurs at step position $\frac{k}{\beta}$, the prefix length is:
\begin{equation}
p_k = \frac{k}{\beta} \times \frac{L}{s} = \frac{k L}{\beta s}
\end{equation}

Then, the computation for a single additional branch is:
\begin{equation}
\begin{aligned}
C_{k} &= \sum_{j=1}^{l}\left(p_{k}+j\right) = l \cdot p_{k} + \frac{l(l+1)}{2} 
\\&= \frac{L}{s} \times \frac{k L}{\beta s} + \frac{L/s (L/s + 1)}{2} \\
&= \frac{k L^{2}}{\beta s^{2}} + \frac{L^{2}/s^{2} + L/s}{2}
\end{aligned}
\end{equation}

Since there are $\alpha\beta s$ branching steps in total, each producing 2 additional branches, the overhead for additional branches is:
\begin{equation}
\begin{aligned}
T_{\text{add}} = \sum_{k=1}^{\beta\alpha s} 2 \times \left[ \frac{k L^{2}}{\beta s^{2}} + \frac{L^{2}/s^{2} + L/s}{2} \right]
\end{aligned}
\end{equation}

Simplifying the first term in Equation 20:
\begin{equation}
\begin{aligned}
\sum_{k=1}^{\beta\alpha s} \frac{2k L^{2}}{\beta s^{2}} 
&= \frac{2 L^{2}}{\beta s^{2}} \times \frac{\beta\alpha s (\beta\alpha s + 1)}{2} \\
&\approx \frac{L^{2}}{\beta s^{2}} \times (\beta\alpha s)^{2} \\
&= \frac{(\beta\alpha)^{2} s^{2} L^{2}}{\beta s^{2}} \\
&= {\beta (\alpha L)^{2}}
\end{aligned}
\end{equation}

Simplifying the second term in Equation 20:
\begin{equation}
\sum_{k=1}^{\beta\alpha s} {L^{2}/s^{2} + L/s} = \beta\alpha s \times ({L^{2}/s^{2} + L/s}) \approx \frac{\beta\alpha L^{2}}{s}
\end{equation}

When $s$ is large, the second term becomes negligible, therefore:
\begin{equation}
T_{\text{add}} = O\left(\beta(\alpha)^{2}L^{2}\right)
\end{equation}

The total cost of PI is:
\begin{equation}
\begin{aligned}
T_{\mathrm{PI}} &= T_{\text{main}} + T_{\text{add}} \\ 
&=
O\left(\alpha^{2}L^{2}\right) + O\left(\beta(\alpha)^{2}L^{2}\right) \\
&= O\left(\alpha^{2}L^{2}\left(1+\beta\right)\right)
\end{aligned}
\end{equation}

Under this setting, we can calculate the computational cost savings ratio of PI compared to the original CoT as:
\begin{equation}
Saving Ratio= 1 - \frac{T_{PI}}{L^{2}} = 1 - \alpha^{2}\left(1 + \beta\right)
\end{equation}

Substituting $\alpha = 0.5$ and $\beta = 0.5$, we obtain:
\begin{equation}
Saving Ratio = 1 - 0.5^{2}\left(1 + 0.5\right)=62.5\%
\end{equation}

Considering that PI only branches during the generation of thinking content and does not branch when presenting conclusions, the compression ratio $\alpha$ is effectively smaller, and the branching positions occur earlier in the sequence. Consequently, the computational overhead is lower than the results calculated under the uniform distribution assumption.

Regarding memory overhead, PI demonstrates significant advantages. The memory overhead analysis can be decomposed into two components: primary memory usage (KV cache) and additional memory overhead (parallel decoding).

First, we analyze the peak memory reduction. In modern LLM inference, the primary memory overhead stems from storing attention keys and values in the KV cache, the size of which scales linearly with the processed sequence length. Standard CoT necessitates storing KV cache for all $L$ tokens, resulting in memory usage of $O(L)$. In contrast, PI effectively reduces the peak sequence length during inference from $L$ to $\alpha L$. Consequently, peak KV cache memory usage decreases from $O(L)$ to $O(\alpha L)$. This memory reduction proves particularly crucial when processing long-context reasoning tasks. Beyond reducing peak memory consumption for individual requests, this optimization enables the system to accommodate a greater number of concurrent requests within the same memory constraints during batch processing, thereby enhancing overall throughput.

Regarding the additional overhead for parallel decoding, modern inference frameworks equipped with prefix caching capabilities (such as vLLM) ensure that our PI method does not triple memory overhead, despite performing two additional forward passes in parallel with the main reasoning branch. When multiple reasoning branches share a common prefix sequence, the corresponding portion of their KV caches needs to be stored only once in physical memory, leveraging technologies such as vLLM's PagedAttention. Consequently, during parallel decoding, the additional branches incur negligible KV cache overhead, as they fully reuse the KV cache already computed by the main reasoning branch.





\section{Human-AI Collaborative Reasoning Interface.}
We have implemented an online version of the Prompt Intervention method for Human-AI Collaborative Reasoning, with the interface shown in Figure 16. By enabling humans to provide real-time guidance on the model's next reasoning action, LRM can achieve reasoning results more efficiently. We hope this interactive experience will assist researchers in better incorporating expert knowledge into the design of static Prompt Intervention.



\begin{figure*}[h!]
  \centerline{\includegraphics[scale=0.31]{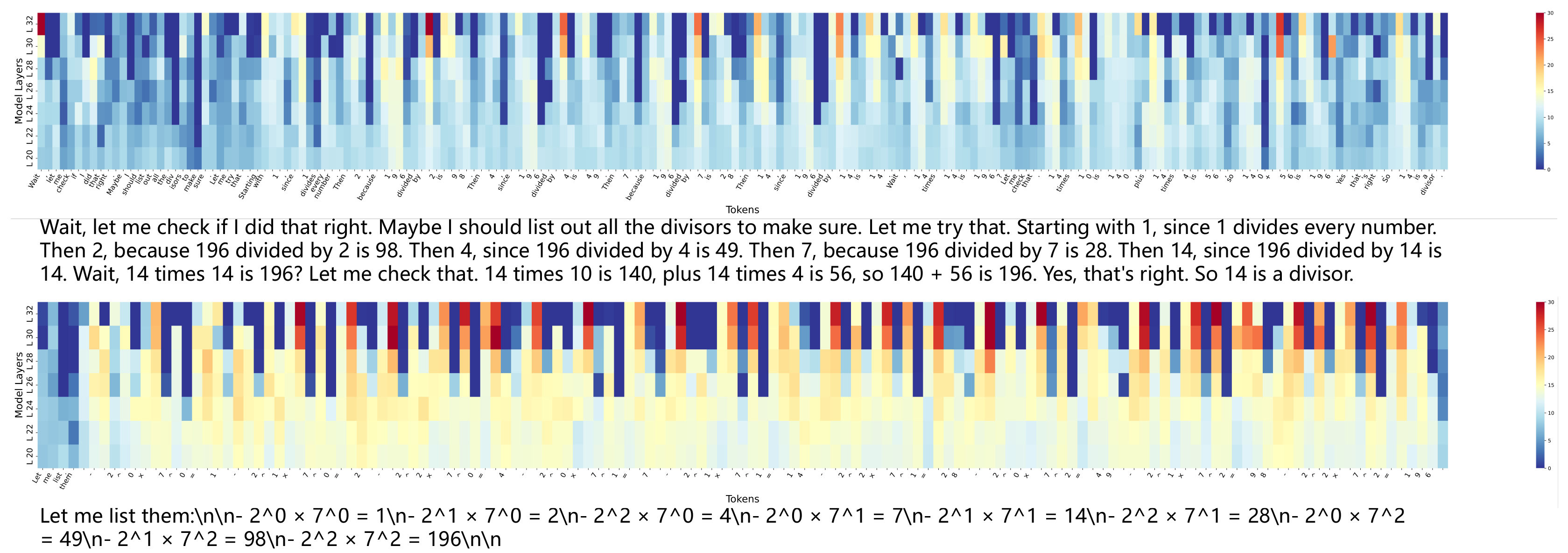}}
  \caption{Comparison of JSD values between early and final layers across different reasoning steps in Qwen3-8B.}
  \label{figure8}
\end{figure*}

{
\setlength{\tabcolsep}{2.5pt}
\begin{table*}[h!]
\centering
\scalebox{0.82}{
\begin{tabular}{@{}lcccccccccccccccccccc@{}} 
\toprule
 \multirow{2}{*}{\textbf{Method}} 
 & \multicolumn{3}{|c}{\textbf{GSM8K}}& \multicolumn{3}{c}{\textbf{MATH-500}} & \multicolumn{3}{c}{\textbf{AMC}} & \multicolumn{3}{c}{\textbf{OlympiadBench}}  & \multicolumn{3}{c}{\textbf{GPQA-D}}  & \multicolumn{3}{c}{\textbf{Minerva}} & \multicolumn{2}{|c}{\textbf{Overall}} \\
   & {Acc$\uparrow$} & {Tok$\downarrow$} & {CR$\downarrow$} & {Acc$\uparrow$} & {Tok$\downarrow$} & {CR$\downarrow$} & Acc$\uparrow$ & Tok$\downarrow$ & {CR$\downarrow$} & {Acc}$\uparrow$  & {Tok$\downarrow$} & {CR$\downarrow$} & {Acc$\uparrow$} & {Tok$\downarrow$} & {CR$\downarrow$} & {Acc$\uparrow$} & {Tok$\downarrow$} & {CR$\downarrow$}& {Acc$\uparrow$} & {CR}$\downarrow$  \\ 
\hline

\textit{Vanilla} & 95.2 & 2191 & 100\% & 92.4 & 5224 & 100\% & 88.8 & 8027 & 100\% & 60.3 & 9414 & 100\% & 58.1 & 9105 & 100\% & \textbf{52.4} & 6850 & 100\%  & \multicolumn{1}{|l}{74.5} & 100\%  \\
\textit{$\pi^d(p,s)$} & \textbf{95.3} & 840 & 38.8\% & \textbf{94.0} & 3,074 & 58.8\% & 89.4 & 4814 & 60.0\% & \textbf{65.5} & 5573 & 59.2\% & {58.6} & 4081 & 44.8\% & 52.0 & 2724 & 39.8\% & \multicolumn{1}{|l}{\textbf{75.8}} &{50.2\%}  \\
\textit{$\pi^d(p,s)'$} & 95.2 & 793 & 36.2\% & {93.2} & 2,862 & 54.8\% & \textbf{90.0} & 4721 & 58.8\% & {64.0} & 5290 & 56.2\% & \textbf{59.2} & 4123 & 45.3\% & 51.6 & 2839 & 41.4\% & \multicolumn{1}{|l}{{75.5}} &{48.8\%}  \\

 \bottomrule
\end{tabular}
}
\caption{Experimental results with different trigger words on Qwen3-8B. For \textit{$\pi^d(p,s)$}, $p$: \textit{'Okay, moving on'}, $s$: \textit{'So, putting it all together'}. For \textit{$\pi^d(p,s)'$}, $p$: \textit{'Continuing along this line of thought'}, $s$: \textit{'Summarizing the existing steps, I'}. The result is statistically significant with $p$-value $<$ 0.05.}
\label{ablation-new}
\end{table*}
}

\begin{figure*}[h!]
  \centerline{\includegraphics[scale=0.15]{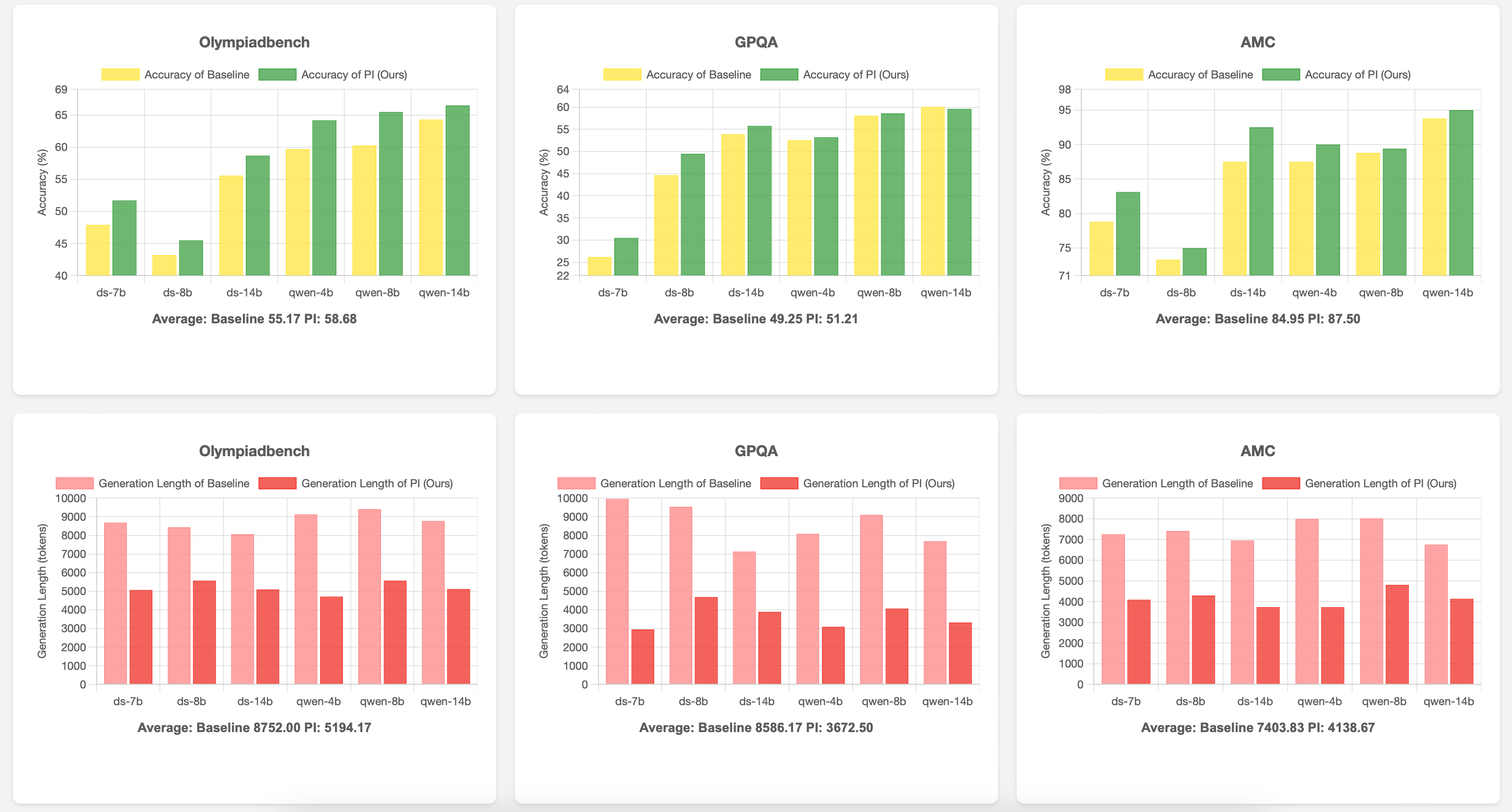}}
  \caption{Performance comparison between PI and the baseline across more models. (ds-7b denotes DeepSeek-R1-Distill-Qwen-7B, ds-8b denotes DeepSeek-R1-Distill-Llama-8B, ds-14b denotes DeepSeek-R1-Distill-Qwen-14B, qwen-4b denotes Qwen3-4B, qwen-8b denotes Qwen3-8B, qwen-14b denotes Qwen3-14B.)}
  \label{figure-bar}
\end{figure*}

\lstdefinelanguage{prompt}{
    numbers=none,
    frame=shadowbox,
    framerule=0.5pt,
    framesep=2pt,
    breaklines=true,
    backgroundcolor=\color{aliceblue},
    basicstyle=\fontsize{9pt}{9pt}\selectfont\ttfamily,
    commentstyle=\color{cyan},
    morecomment=[l]{//},
    moredelim=[is][\color{frenchrose}\bfseries]{<<<}{>>>},
    moredelim=[is][\color{awesome}\bfseries]{***}{***},
    moredelim=[is][\color{azure(colorwheel)}\bfseries]{///}{///},
    moredelim=[is][\color{coquelicot}\bfseries]{|||}{|||},
}\begin{figure*}[!htb]\begin{lstlisting}[language=prompt]
|||Question|||: What is the smallest positive perfect cube that can be written as the sum 
of three consecutive integers?
|||Answer|||: 
///Step 1///: <think>\nOkay, so I need to find the smallest positive perfect cube that 
can be written as the sum of three consecutive integers. Hmm, let's start by 
understanding the problem.
///Step 2///: First, a perfect cube is a number that's an integer raised to the power of
three, like 1, 8, 27, 64, etc. The question is asking for the smallest such number that 
can also be expressed as the sum of three consecutive integers.
///Step 3///: Let me think about how to represent three consecutive integers. Let's say 
the three consecutive integers are n-1, n, and n+1. Wait, but maybe it's easier to let 
the middle number be n, so the three numbers would be n-1, n, n+1. Then their sum would 
be (n-1) + n + (n+1). Let me compute that.
///Step 4///: ***Okay, moving on.*** Adding them up: (n - 1) + n + (n + 1) = 3n. Oh, 
that's interesting. So the sum of three consecutive integers is always three times the 
middle number. So, if I want this sum to be a perfect cube, 3n must be a perfect cube.
///Step 5///: So, 3n = k^3, where k is a positive integer. Then n = k^3 / 3. Since n has 
to be an integer (because we're talking about integers), k^3 must be divisible by 3. That 
means k must be divisible by 3. Let me check that. If k is divisible by 3, then k = 3m 
for some integer m. Then k^3 = (3m)^3 = 27m^3. So n = 27m^3 / 3 = 9m^3. 
///Step 6///: ***So, putting it all together***, the sum of three consecutive integers is 
3n = 3*(9m^3) = 27m^3. Therefore, the sum is 27m^3, which is a perfect cube since 27 is 
3^3. Therefore, the sum is (3m)^3. 

...
...
***<Omitted Steps>***
...
...

///Step 15///:
***Top-4 tokens: ['Another', 'Therefore', 'So', '!']***
***Top-4 probs: [0.57, 0.24, 0.19, 0.0]***
***High Entropy***
///Original Content///: Another way to think about it: Let me try small perfect cubes and
see if they can be written as the sum of three consecutive integers.///("Exploration")///
reasoning_score: 0.272
sequence_prob: 0.766
///"Progression" Intervention Content///: Okay, moving on. I think that's solid. So the 
answer is 27.\n</think>
reasoning_score: ***0.649***
sequence_prob: ***0.949***
***[Chosen]: max score***
///"Summary" Intervention Content///: So, putting it all together, the smallest positive
perfect cube is 27.\n</think>
reasoning_score: 0.643
sequence_prob: 0.931
\end{lstlisting}
\caption{Example showing the specific process of one PI intervention.}
\label{detailed}
\end{figure*}

\lstdefinelanguage{prompt}{
    numbers=none,
    frame=shadowbox,
    framerule=0.5pt,
    framesep=2pt,
    breaklines=true,
    backgroundcolor=\color{aliceblue},
    basicstyle=\fontsize{9pt}{9pt}\selectfont\ttfamily,
    commentstyle=\color{cyan},
    morecomment=[l]{//},
    moredelim=[is][\color{frenchrose}\bfseries]{<<<}{>>>},
    moredelim=[is][\color{awesome}\bfseries]{***}{***},
    moredelim=[is][\color{azure(colorwheel)}\bfseries]{///}{///},
    moredelim=[is][\color{coquelicot}\bfseries]{|||}{|||},
}\begin{figure*}[!htb]\begin{lstlisting}[language=prompt]
|||Question|||: A regular hexagon can be divided into six equilateral triangles. If the
perimeter of one of the triangles is 21 inches, what is the perimeter, in inches, of the
regular hexagon?
|||Answer|||: 
///Step 1///: <think>\nOkay, let's see. The problem says a regular hexagon can be divided
into six equilateral triangles. If the perimeter of one of the triangles is 21 inches, 
what is the perimeter of the hexagon?\n\n
///Step 2///: First, I need to visualize a regular hexagon. A regular hexagon has six 
equal sides and six equal angles. When it's divided into six equilateral triangles, I 
imagine drawing lines from the center of the hexagon to each of the six vertices. That 
would create six congruent triangles, right? Each of these triangles should be 
equilateral because all sides are equal in a regular hexagon.\n\n
///Current Step///:
***Top-4 tokens: ['So', '#', '!', '"']***
***Top-4 probs: [1.0, 0.0, 0.0, 0.0]***
***Entropy:0***
///Original Content///: So, each of these triangles is equilateral. That means all three
sides of each triangle are the same length. The perimeter of one triangle is given as 21 
inches. Since an equilateral triangle has three equal sides, each side length would be 21 
divided by 3. Let me calculate that: 21 / 3 = 7 inches. So each side of the triangle is 7 
inches. 
***[Chosen]***
///"Summary" Intervention Content///: So, putting it all together. I need to relate the 
perimeter of one of these triangles to the perimeter of the hexagon. Let me think. Each 
triangle has three sides. But wait, in the hexagon, each side is actually a side of one 
of these triangles. Wait, but when you divide the hexagon into triangles by connecting 
the center to the vertices, each triangle has two sides that are radii of the hexagon and 
one side that is a side of the hexagon. Wait, but the problem says the triangles are 
equilateral. So if the triangles are equilateral, then all three sides must be equal. 
That would mean that the sides of the hexagon are equal to the radii of the hexagon. Hmm, 
maybe I need to recall some properties of regular hexagons.\n\n 
***[Rejected] (Low Quality)***
\end{lstlisting}
\caption{Examples where forced low-entropy intervention leads to low-quality steps.}
\label{high-e}
\end{figure*}

\lstdefinelanguage{prompt}{
    numbers=none,
    frame=shadowbox,
    framerule=0.5pt,
    framesep=2pt,
    breaklines=true,
    backgroundcolor=\color{aliceblue},
    basicstyle=\fontsize{9pt}{9pt}\selectfont\ttfamily,
    commentstyle=\color{cyan},
    morecomment=[l]{//},
    moredelim=[is][\color{frenchrose}\bfseries]{<<<}{>>>},
    moredelim=[is][\color{awesome}\bfseries]{***}{***},
    moredelim=[is][\color{azure(colorwheel)}\bfseries]{///}{///},
    moredelim=[is][\color{coquelicot}\bfseries]{|||}{|||},
}\begin{figure*}[!htb]\begin{lstlisting}[language=prompt]
Given a question, you need to generate a seemingly relevant but actually inconsequential
statement (No-Op) that can be added to the question without affecting its solution.
Generate a No-Op statement that:
1. Is short and concise
2. Seems relevant to the context
3. Is natural and fits grammatically
4. Contains misleading numbers, but do not affect the final outcome.
You could refer to the following example:
***[Example]***
|||Question|||: Liam wants to buy some school supplies. He buys 24 erasers that now cost
$6.75 each, 10 notebooks that now cost $11.0 each, and a ream of bond paper that
now costs $19. How much should Liam pay now?
|||No-Op statement|||: Assuming that due to inflation, prices were 10% cheaper last year.
|||Explanation|||: The question asks about the current price, while the No-Op statement provides
a condition about last year\s lower price, which does not affect the current calculation. 
Therefore, it is misleading but does not impact the solution. 
Generate a No-Op statement for the following questions by mimicking the above example, 
without providing |||Explanation|||: 
|||Question|||: {}
|||No-Op statement|||: 
\end{lstlisting}
\caption{Instruction for GSM-NoOp Construction (part 1).}
\label{noop_template1}
\end{figure*}

\lstdefinelanguage{prompt}{
    numbers=none,
    frame=shadowbox,
    framerule=0.5pt,
    framesep=2pt,
    breaklines=true,
    backgroundcolor=\color{aliceblue},
    basicstyle=\fontsize{9pt}{9pt}\selectfont\ttfamily,
    commentstyle=\color{cyan},
    morecomment=[l]{//},
    moredelim=[is][\color{frenchrose}\bfseries]{<<<}{>>>},
    moredelim=[is][\color{awesome}\bfseries]{***}{***},
    moredelim=[is][\color{azure(colorwheel)}\bfseries]{///}{///},
    moredelim=[is][\color{coquelicot}\bfseries]{|||}{|||},
}\begin{figure*}[!htb]\begin{lstlisting}[language=prompt]
Please combine the following math question and No-Op phrase into a single, natural-
sounding question. The No-Op phrase should be integrated smoothly without changing the 
mathematical meaning.
|||Math Question|||: {}
|||No-Op Phrase|||: {}
|||Combined Question|||:
\end{lstlisting}
\caption{Instruction for GSM-NoOp Construction (part 2).}
\label{noop_template2}
\end{figure*}

\lstdefinelanguage{prompt}{
    numbers=none,
    frame=shadowbox,
    framerule=0.5pt,
    framesep=2pt,
    breaklines=true,
    backgroundcolor=\color{aliceblue},
    basicstyle=\fontsize{9pt}{9pt}\selectfont\ttfamily,
    commentstyle=\color{cyan},
    morecomment=[l]{//},
    moredelim=[is][\color{frenchrose}\bfseries]{<<<}{>>>},
    moredelim=[is][\color{awesome}\bfseries]{***}{***},
    moredelim=[is][\color{azure(colorwheel)}\bfseries]{///}{///},
    moredelim=[is][\color{coquelicot}\bfseries]{|||}{|||},
}\begin{figure*}[!htb]\begin{lstlisting}[language=prompt]
|||Question|||: Simplify $\\sqrt{242}$.
///Original Response///: Okay, so I need to simplify the square root of 242. Hmm, let me
think. Simplifying square roots usually involves factoring the number under the root into
perfect squares and other numbers. Let me recall, a perfect square is a number that's the 
square of an integer, like 4 (which is 2^2), 9 (3^2), 16 (4^2), and so on. So, if I can 
break down 242 into a product of a perfect square and another number, then I can take the
square root of the perfect square out of the radical.\n\nFirst, let me try dividing 242 
by some small perfect squares. Let me start with 2. 242 divided by 2 is 121. Wait, 121 
sounds familiar. Isn't that a perfect square? Yes! 11 squared is 121. So that means 242
can be written as 2 times 121. Therefore, sqrt242 = sqrt(2 x 121). \n\nNow, since 121 is a
perfect square, I can separate the square root into the product of the square roots. So, 
sqrt(2 x 121) = sqrt2 x sqrt121. And since sqrt121 is 11, this simplifies to 11sqrt2.
\n\nWait, let me check if there's a larger perfect square factor that I might have missed.
Let me try dividing 242 by 4. 242 divided by 4 is 60.5, which isn't an integer, so that's 
not helpful. How about 9? 242 divided by 9 is approximately 26.888... Not an integer. 16? 
242 divided by 16 is 15.125. Still not an integer. 25? 242 divided by 25 is 9.68. Nope. 
36? 242 divided by 36 is about 6.722... Not an integer. 49? 242 divided by 49 is roughly 
4.938... Not an integer. 64? 242 divided by 64 is 3.78125. Still no. 81? 242 divided by 81
is about 2.987... Not an integer. 100? 242 divided by 100 is 2.42. Not an integer. So, the
only perfect square that divides 242 evenly is 121, which is 11 squared. \n\nTherefore, my
initial simplification seems correct. So sqrt242 simplifies to 11sqrt2. Let me verify this 
by squaring 11sqrt2 to see if I get back 242. \n\nCalculating (11sqrt2)^2: 11 squared is 
121, and (sqrt2) squared is 2. Multiplying those together gives 121 x 2 = 242. Perfect, 
that matches the original number under the square root. \n\nSo, I think that's it. The
simplified form of sqrt242 is 11sqrt2. Let me just check if there's any other way to 
factor 242 that I might have missed. For example, maybe breaking it down into more than 
one perfect square? Let me see. 242 factors into 2 x 121, and 121 is 11^2. 
There's no other way to factor 242 into perfect squares, right? Because 2 is a prime 
number, so it can't be broken down further into smaller integers. Therefore, the only 
perfect square factor is 121. \n\nAlternatively, if I factor 242 completely into primes,
maybe that would help. Let me try that. Starting with 242. Divided by 2 gives 121. Then
121 is 11 x 11. So the prime factorization of 242 is 2 x 11 x 11. So, written as 2 x 11^2.
Therefore, sqrt(2 x 11^2) = sqrt2 x sqrt11^2 = 11sqrt2. Yep, same result. \n\nSo, I think 
I've confirmed it multiple ways. The simplified form is 11sqrt2. I don't think there's any
more simplification possible because 2 is a prime number and doesn't have any square 
factors other than 1. Therefore, the answer should be 11sqrt2.\n\n**Final Answer**\nThe 
simplified form of $\\sqrt{242}$ is \\boxed{11\\sqrt{2}}.\n***</think>***\n\nTo simplify the 
square root of 242, we begin by factoring the number under the square root into its prime 
components or into a product that includes a perfect square.\n\n---\n\n
Step 1: Prime Factorization\n\nWe start by factoring 242:\n\n$$\n242 \\div 2 = 
121\n$$\n\nNow, 121 is a well-known perfect square:\n\n$$\n121 = 11^2\n$$\n\nSo, the full
prime factorization of 242 is:\n\n$$\n242 = 2 \\times 11^2\n$$\n\n---\n\n 
Step 2: Apply the Square Root\n\nUsing the property of square roots:\n\n$$\n
\\sqrt{a \\times b} = \\sqrt{a} \\times \\sqrt{b}\n$$\n\nWe can write:\n\n$$\n
\\sqrt{242} = \\sqrt{2 \\times 11^2} = \\sqrt{2} \\times \\sqrt{11^2}\n$$\n\n
$$\n= \\sqrt{2} \\times 11 = 11\\sqrt{2}\n$$\n\n---\n\n 
Step 3: Final Simplification\n\nSince 2 is a prime number and has no square factors other
than 1, the expression $11\\sqrt{2}$ is fully simplified.\n\n---\n\n 
***Final Answer\n\\boxed{11\\sqrt{2}}\n***
///PI's Response///: Okay, so I need to simplify the square root of 242. Let me think
about how to approach this. I remember that simplifying square roots involves factoring 
the number under the root into perfect squares and other factors. The goal is to take out 
any perfect squares from under the square root sign. ***Okay, Moving on.*** I should factor 242 
into its prime factors. Let me start by dividing by the smallest prime numbers. 242 is 
even, so it's divisible by 2. Dividing 242 by 2 gives me 121. Now, 121 seems familiar. 
Isn't that a perfect square? Yes, 11 times 11 is 121. So, breaking it down, 242 is 2 
multiplied by 11 squared. So, writing that out, 242 = 2 x 11^2. Now, when simplifying 
the square root, I can separate the square factor. The square root of a product is the 
product of the square roots. So, sqrt(2 x 11^2) can be written as sqrt2 x sqrt(11^2). 
 The square root of 11 squared is just 11, right? Because the square root and the 
square cancel each other out. So that simplifies to 11sqrt2. ***Okay, Moving on.*** I think 
that's as simplified as it gets. ***So, putting it all together,*** the simplified form is 
indeed 11sqrt2. I don't think I missed any steps here. No other perfect squares are 
present in the factors, so this should be the simplest form. ***</think>*** To simplify 
$\\sqrt{242}$, we start by factoring 242 into its prime components. \n\n242 is even, so it
is divisible by 2:  \n$$\n242 = 2 \\times 121\n$$  \n121 is a perfect square ($11^2$), so 
we rewrite:  \n$$\n242 = 2 \\times 11^2\n$$  \n\nUsing the property of square roots, 
$\\sqrt{a \\times b} = \\sqrt{a} \\times \\sqrt{b}$, we separate the square factor:  
\n$$\n\\sqrt{242} = \\sqrt{2 \\times 11^2} = \\sqrt{2} \\times \\sqrt{11^2}\n$$  \n\nSince 
$\\sqrt{11^2} = 11$, this simplifies to:  \n$$\n\\sqrt{242} = 11\\sqrt{2}\n$$  \n\nNo 
further simplification is possible because 2 is a prime number with no square factors.
\n\n***$$\n\\boxed{11\\sqrt{2}}\n$$***

\end{lstlisting}
\caption{More comparison examples of PI versus vanilla CoT.}
\label{case_study_more_1}
\end{figure*}

\lstdefinelanguage{prompt}{
    numbers=none,
    frame=shadowbox,
    framerule=0.5pt,
    framesep=2pt,
    breaklines=true,
    backgroundcolor=\color{aliceblue},
    basicstyle=\fontsize{9pt}{9pt}\selectfont\ttfamily,
    commentstyle=\color{cyan},
    morecomment=[l]{//},
    moredelim=[is][\color{frenchrose}\bfseries]{<<<}{>>>},
    moredelim=[is][\color{awesome}\bfseries]{***}{***},
    moredelim=[is][\color{azure(colorwheel)}\bfseries]{///}{///},
    moredelim=[is][\color{coquelicot}\bfseries]{|||}{|||},
}\begin{figure*}[!htb]\begin{lstlisting}[language=prompt]
|||Question|||: We know that visionaries in California got all their ideas from taking what?
A. Visionaries in California got their ideas from different sources. 
B. I have no comment. 
C. There is no single source from which visionaries in California got their ideas. 
D. Visionaries in California got their ideas from drugs.
E. Visionaries in California got their ideas from dreams.
F. Visionaries in California got their ideas from meditation.
G. Visionaries in California got their ideas from LSD.
///Original Response///: Okay, let's try to figure out this question. The question is asking:
"We know that visionaries in California got all their ideas from taking what?" And there 
are several options provided.
First, I need to understand what the question is really about. It seems to be a play on
words, maybe a riddle or a pun. The key here is that it's asking what they "got all their
ideas from taking." The word "taking" is important, it suggests something they consume or 
partake in, like a substance.
Looking at the options, most of them are about sources of ideas, like different sources, 
meditation, LSD, etc. Option A says they got ideas from different sources, which is a bit
vague. Option D mentions drugs, which is more specific. Option G specifically mentions 
LSD, which is a hallucinogenic drug. Option E talks about dreams, which is another source
but not necessarily something they "take".
Now, considering the context, when we think of visionaries in California, especially 
during the 60s and 70s, LSD was a significant part of the counterculture movement. Many 
people in that era used LSD for its perceived mind-expanding effects, which could 
influence creativity and visionary ideas.
So, the question is likely a riddle where "taking" refers to consuming LSD. Therefore, the
answer is probably G, Visionaries in California got their ideas from LSD. ***</think>***
The question is a riddle where "taking" refers to consuming a substance. Considering the 
context of California's counterculture movement, LSD is linked to creativity and visionary
ideas. ***\boxed{G}*** ***(incorrect)***
///PI's Response///: Okay, so I have this question here asking about where visionaries in 
California got their ideas from, and the options include things like drugs, LSD, 
meditation, dreams, and others. Let me break this down step by step.First, I need to 
understand the context. The question is about visionaries in California. I'm thinking of 
people like Steve Jobs, maybe some of the early tech pioneers from Silicon Valley, or 
perhaps even people in the counterculture movements like the 60s and 70s. Those were times
when a lot of new ideas and movements started, so that's probably the timeframe we're 
looking at. Now, the options given are A to G, each suggesting a different source of 
ideas. Let's go through them one by one. The options are: A. Visionaries got ideas from 
different sources. B. I have no comment. C. There's no single source. D. Drugs. E. Dreams.
F. Meditation. G. LSD specifically. ***Okay, moving on.*** I know that in the 60s and 70s, 
California, especially places like San Francisco and Silicon Valley, were hubs for various
movements. The counterculture movement was big then, which included a lot of 
experimentation with psychedelics. I remember reading that LSD was a significant part of 
that. People like Timothy Leary were promoting it, and it's often linked with the idea of 
expanding consciousness and creativity. ***Wait, let me verify.*** Steve Jobs was known to have
taken LSD, and he talked about it influencing his thinking. Similarly, there are stories 
about other tech figures who were part of the counterculture movement using psychedelics 
as a way to explore new ideas. So that would support option D and G. ***Okay, moving on.*** I 
also know that meditation and mindfulness practices were becoming more popular in 
California during that time. It's part of the New Age movement, which was influenced by 
eastern philosophies. So option F might also be a factor, but I think it's more about 
personal growth than directly generating ideas. As for dreams, I'm not sure there's a 
strong link between visionary ideas and specifically recalling dreams. Maybe some people 
do get ideas from dreams, but it's not as widespread as the use of psychedelics. So option
E might not be the best fit. ***So, putting it all together***, the question is asking what they
got their ideas from. Since there's a well-documented connection between the 
counterculture movement in California and the use of LSD and other psychedelics for 
inspiration, options D and G both seem relevant. But since LSD is a specific type of drug,
maybe G is more precise. Wait, but the options include A and C, which suggest that there's
no single source. Maybe the answer is C because while LSD was a common factor, not 
everyone used it, and people drew from different sources. Or it's D because it's more 
general. ***Okay, moving on.*** I need to choose between A, C, D, E, F, G, and B. B is just a 
refusal to comment, so probably not. The correct answer is likely either D or G, but maybe 
C is more accurate because it's acknowledging that there's no one single source. However, 
if the question is looking for the most common or significant factor, then D or G might be 
better. ***So, putting it all together***, the most accurate answer is probably C, that there's 
no single source, because while LSD was influential, many visionaries drew from various 
experiences and contexts, not just drugs. Therefore, C would be the best choice as it 
acknowledges the diversity of sources rather than pointing to one specific thing. ***</think>***
The question asks about the source of ideas for visionaries in California, with options 
including drugs, LSD, meditation, and others. Considering the context of the 60s and 70s 
counterculture movement, LSD was a significant influence, but it wasn't the only source.
Visionaries drew from diverse experiences, making the most accurate answer that there's no 
single source. ***\\boxed{C}*** ***(correct)***
\end{lstlisting}
\caption{More comparison examples of PI versus vanilla CoT.}
\label{case_study_more_2}
\end{figure*}

\begin{figure*}[h!]
  \centerline{\includegraphics[scale=0.55]{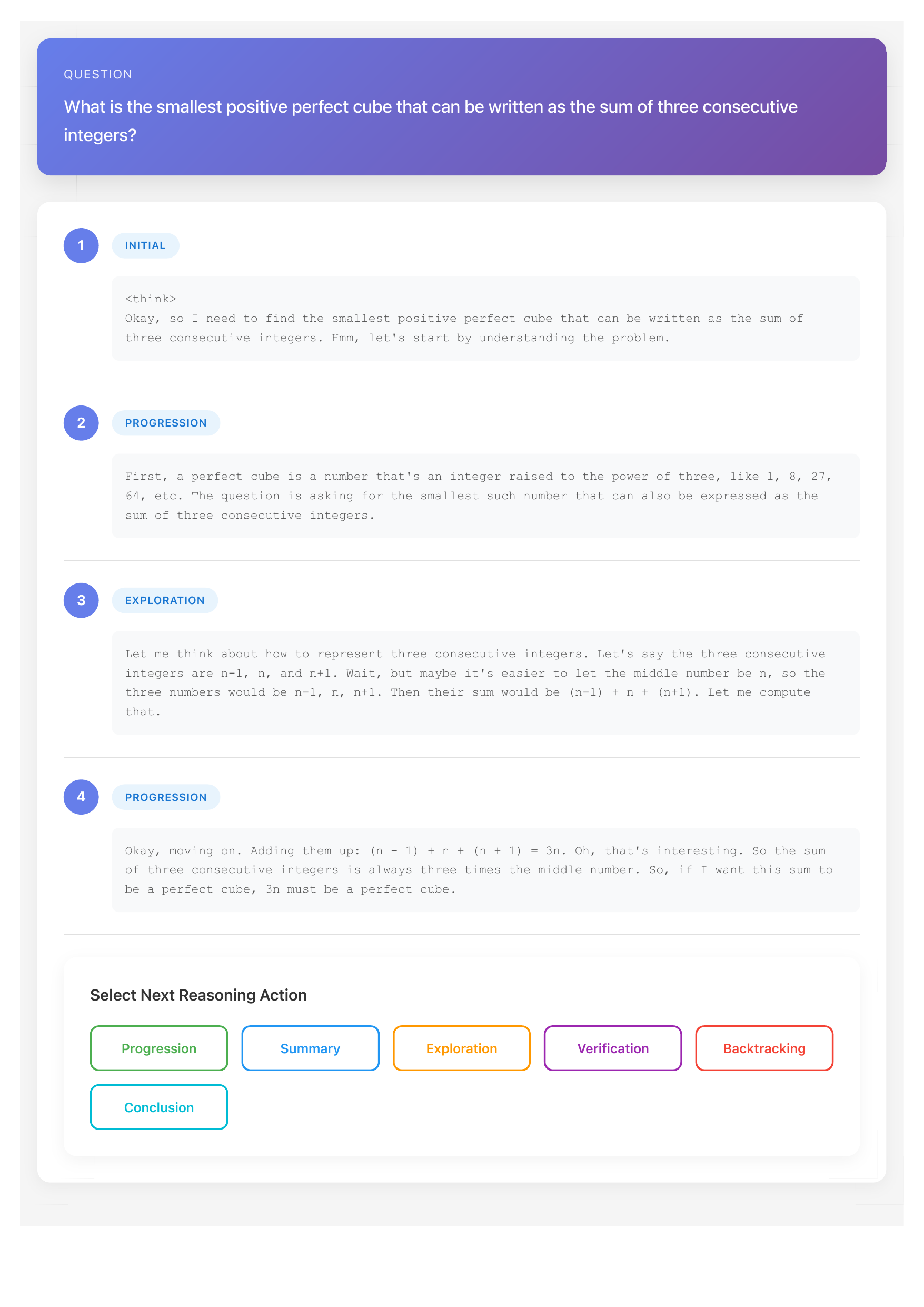}}
  \caption{Demonstration of human-AI collaborative reasoning interface.}
  \label{figure15}
\end{figure*}

\end{document}